\documentclass[lettersize,journal]{IEEEtran}
\usepackage{amsmath,amsfonts}
\usepackage{algorithmic}
\usepackage{algorithm}
\usepackage{array}
\usepackage[caption=false,font=normalsize,labelfont=sf,textfont=sf]{subfig}
\usepackage{textcomp}
\usepackage{stfloats}
\usepackage{url}
\usepackage{verbatim}
\usepackage{graphicx}
\usepackage{multirow}
\usepackage{booktabs}
\usepackage{tabularx}
\usepackage{threeparttable}
\usepackage[T1]{fontenc}
\usepackage{lscape}
\usepackage{cite}
\usepackage{pifont} 
\usepackage{stackengine}
\usepackage[hidelinks, colorlinks]{hyperref}
\usepackage{xcolor}
\newcommand{\cmark}{\ding{51}}  
\newcommand{\xmark}{\ding{55}}  
\begin{document}

\title{SRMF: A Data Augmentation and Multimodal Fusion Approach for Long-Tail UHR Satellite Image Segmentation}

\author{Yulong~Guo\textdagger, Zilun Zhang\textdagger, Yongheng Shang, Tiancheng Zhao, Shuiguang Deng, Yingchun Yang, Jianwei Yin
\thanks{
\textdagger:The first two authors contributed equally to this work. }
\thanks{\textit{Corresponding author: Yingchun Yang}. }
\thanks{
Yulong Guo, Zilun Zhang, Yingchun Yang, Jianwei Yin and Shuiguang Deng are with the College of Computer Science and Technology, Zhejiang University, Hangzhou, China; (e-mail: luogu1024116558@gmail.com; zilun.zhang@mail.zju.edu.cn; yyc@cs.zju.edu.cn; zjuyjw@zju.edu.cn; dengsg@zju.edu.cn).
}
\thanks{Yongheng Shang is with the Haina Institute of Zhejiang University, Advanced Technology Institute, Zhejiang University. (e-mail: yh\_shang@zju.edu.cn)}
\thanks{Tiancheng Zhao is with the Binjiang Research Institute of Zhejiang University. (e-mail: tianchez@zju-bj.com)}
\thanks{This study received funding from the National Key Research and Development Program of China under grant number 2023YFD2000101 and the Hainan Province Science and Technology Special Fund of the Hainan Provincial Department of Science and Technology (ZDYF2022SHFZ323).}
}

\markboth{Journal of \LaTeX\ Class Files,~Vol.~14, No.~8, August~2021}%
{Shell \MakeLowercase{\textit{et al.}}: A Sample Article Using IEEEtran.cls for IEEE Journals}


\maketitle

\begin{abstract}
The long-tail problem presents a significant challenge to the advancement of semantic segmentation in ultra-high-resolution (UHR) satellite imagery. While previous efforts in UHR semantic segmentation have largely focused on multi-branch network architectures that emphasize multi-scale feature extraction and fusion, they have often overlooked the importance of addressing the long-tail issue. In contrast to prior UHR methods that focused on independent feature extraction, we emphasize data augmentation and multimodal feature fusion to alleviate the long-tail problem. In this paper, we introduce SRMF, a novel framework for semantic segmentation in UHR satellite imagery. Our approach addresses the long-tail class distribution by incorporating a multi-scale cropping technique alongside a data augmentation strategy based on semantic reordering and resampling. To further enhance model performance, we propose a multimodal fusion-based general representation knowledge injection method, which, for the first time, fuses text and visual features without the need for individual region text descriptions, extracting more robust features. Extensive experiments on the URUR, GID, and FBP datasets demonstrate that our method improves mIoU by 3.33\%, 0.66\%, and 0.98\%, respectively, achieving state-of-the-art performance. Code is available at: https://github.com/BinSpa/SRMF.git.
\end{abstract}


\section{Introduction}
\IEEEPARstart{W}{ith} the rapid development of remote sensing satellites, a large volume of ultra-high-resolution(UHR) satellite images are being collected and processed. These images are utilized to extract various geospatial objects such as roads \cite{lian2020road}, water bodies \cite{Glh-water}, buildings \cite{brrnet} \cite{luo2024multimodal}, and forests \cite{forest}. They are also  employed to study the changes in land use and land cover(LULC) categories \cite{lclu}. In order to fit the aforementioned applications, a multitude of UHR satellite image processing algorithms have been proposed, such as semantic segmentation, style transfer, image restoration and image super resolution. The processing of UHR remote sensing satellite imagery is playing an increasingly significant role in fields such as disaster management \cite{shafapourtehrany2023comprehensive}, environmental monitoring \cite{himeur2022using} \cite{li2020review}, land resource protection \cite{radovcaj2020global}, urban planning \cite{lin2021identifying}, and ecological construction \cite{zhu2021detecting}.

There has been a plethora of outstanding work in the field of image segmentation within the realm of remote sensing\cite{iterdanet, bifdanet, dbblendmask}. As one of the most applied technologies for UHR image processing, semantic segmentation requires pixel-level classification of the images. Along with the continuous advancement of deep learning, models including\cite{deeplabv3p, fpn, pspnet, unet, unet++, segformer,pvtv2,strudel2021segmenter,vit,swintrans} have made significant progress in the field of semantic segmentation. The prevailing approach involves designing a multi-branch network to capture features at different scales, followed by feature fusion to enhance the performance of semantic segmentation, including GLNet\cite{GLNet}, Wiconet\cite{wconet}, ICNet\cite{icnet} and FctlNet\cite{Fctlnet}. However, the aforementioned methods, when extracting features from UHR satellite images, are based on randomly cropped local images and do not consider the integrity of the ground objects. Another approach focuses on constructing a processing pipeline for the entire image, as demonstrated in studies such as ISDNet\cite{isdnet}, EHSNet\cite{ehsnet}, WSDNet\cite{urur} and GPWFormer\cite{GPWFormer}. Given the importance of selecting the content and dimensions of training images, GRNet\cite{patchproposal} and GeoAgent\cite{GeoAgent} each developed specialized modules to choose images for input into neural networks. However, current research on semantic segmentation methods for UHR satellite image remains focused on the extraction and fusion of features from geospatial objects of varying sizes. Nevertheless, These methods do not take into account the impact of the long-tail problem.

\begin{figure}[htbp] 
\centering
\includegraphics[width=\linewidth]{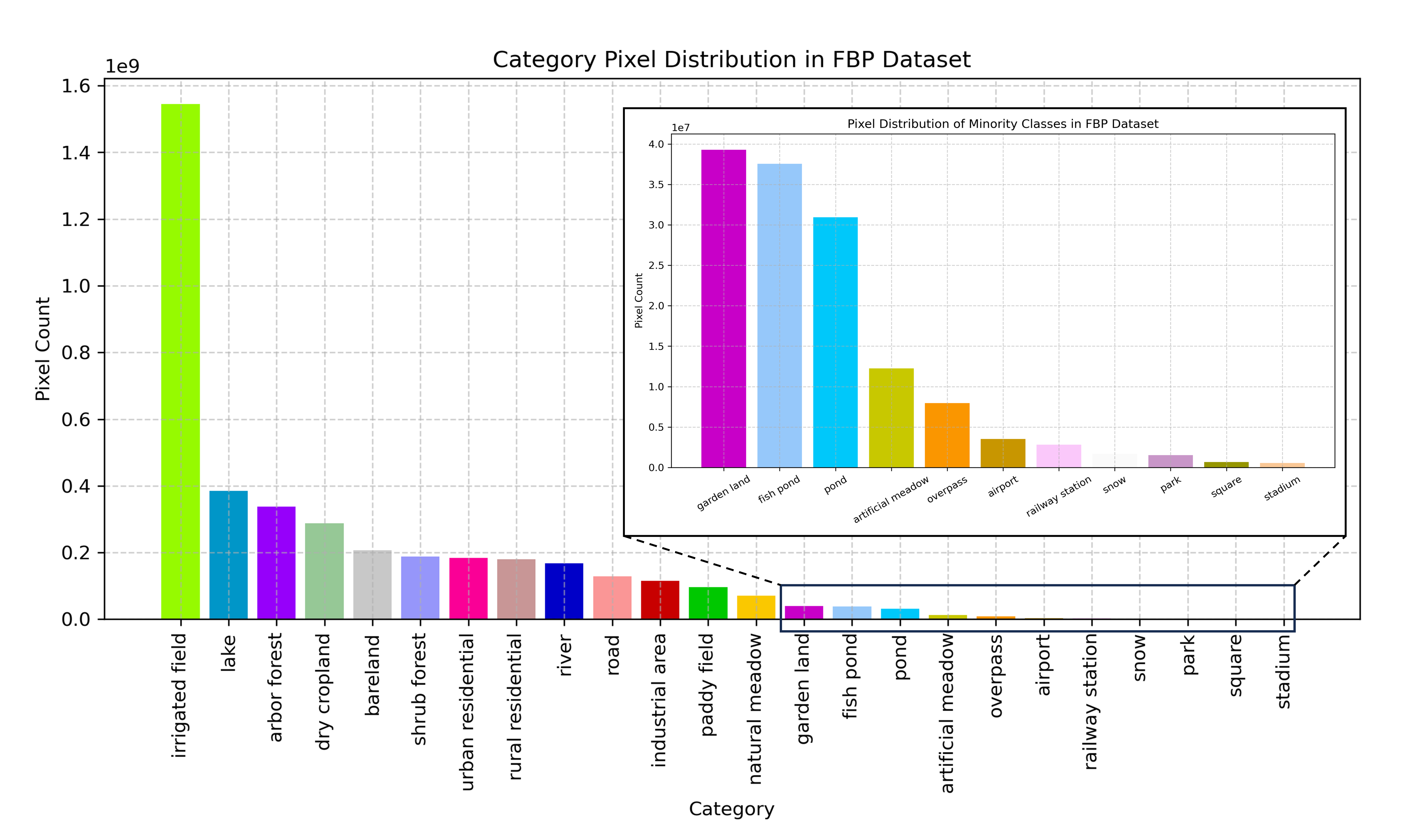} 
\caption{The distribution of class pixel counts in the training set of FBP dataset.}
\label{fig:pixel_dist}
\end{figure}

Due to the differences between natural images and UHR remote sensing images, which commonly cover wide areas of ground with a diverse range of massive ground objects like rivers, farmlands, mountains, and smaller objects such as buildings, ponds, and parks, these characteristics lead to a serious processing problem known as class imbalance, further manifesting as a pronounced long-tail distribution. As depicted in Figure \ref{fig:pixel_dist}, we present the pixel distribution of training classes for the FBP\cite{FBP} remote sensing satellite image datasets,  which exhibit a characteristic long-tail distribution. As described in \cite{ssrs_survey}, unlike natural image domains, this distribution is inherent to satellite imagery. In particular, in urban regions, buildings dominate naturally, whereas natural vegetation such as shrubs, grasslands, and forests occupy expansive areas in mountainous regions. In contrast, features such as airports, stations, and parks are inevitably in the minority. 

The study of \cite{lt-93} has demonstrated that in long-tail distributions, the majority classes predominantly influence the model's parameter updates, leading to sub-optimal performance for minority classes. This issue is particularly pronounced in UHR satellite imagery, where the long-tailed distribution exacerbates the challenges posed by high intraclass variance and low interclass variance. Specifically, categories like meadow and building exhibit significant intra-class variance. For instance, meadow encompasses both artificial and natural types, while building includes residential, industrial, and station structures. These categories require ample training samples to enable models to learn robust features. However, being tail classes, they often lack sufficient training data. Moreover, random cropping during semantic segmentation training may not adequately capture these categories, further deteriorating their performance. In contrast, categories such as river, pond, and lake demonstrate low variance between classes. Models tend to confuse these with similar categories, for example, barren land with parks or gardens, and rivers or ponds with lakes. To address this, establishing a representative center for each category can guide the model's learning process. Therefore, this work focuses on designing data augmentation methods to balance the number of training samples across categories and explores the injection of external knowledge to serve as feature space centers, effectively guiding model training.

Multi-scale cropping plays a crucial role in the UHR satellite images semantic segmentation methods. When viewed from the perspective of long-tail distribution, simple multi-scale cropping does not alter the inherent data distribution. Consequently, dominant categories such as rivers, buildings, and farmland continue to be frequently sampled, while minority categories like ponds, lawns, and greenhouses remain challenging to capture. The larger the area occupied by a land cover object, the stronger its spatial locality, and vice versa for objects confined to smaller areas. For instance, the periphery of farmland is likely to be more farmland, and the surroundings of buildings are probably other architectural complexes. However, the vicinity of ponds, parks, and grasslands may no longer consist of the same features, instead, it might be taken over by areas like rivers or buildings. Thus, the selection of sampling regions is crucial for the task of UHR image semantic segmentation. The most commonly utilized cropping methods, such as those employed in FctlNet\cite{Fctlnet}, involve center-region cropping, which selects an anchor area and samples several larger surrounding regions based on this central area as training samples. This approach tends to exacerbate the local specificity issue mentioned previously. Consequently, we have refined the cropping method by expanding it into a \textbf{Multi-Scale Anchored Region Sampling(MARS)} method. This innovation positions the anchor area at varying locations within differently scaled cropped regions, thereby enhancing the diversity of sampling.

Due to the long-tail problem present in remote sensing satellite images, the scene variations introduced by multi-scale cropping alone are insufficient to alter the inherent class distribution in the training data. Although many studies, such as \cite{lt-nc} and \cite{lt-ss}, have optimized loss functions to address the long-tail issue, the performance gains remain limited. In classification tasks, sampling strategies for training data have been extensively studied. Many efforts have been made to balance the dataset distribution through resampling methods, such as oversampling minority classes\cite{oversample}, undersampling majority classes\cite{undersample}, and resampling based on class frequency\cite{balancesample1}\cite{balancesample2}. While these approaches have effectively improved performance on long-tail data, they are primarily designed for classification or object detection tasks, and may not be directly applicable to the UHR satellite image semantic segmentation scenario. Currently, no training data balancing methods have been specifically designed to address the long-tail problem in UHR satellite images. Howerver, in recent years, instance segmentation models centered around SAM\cite{sam} have demonstrated remarkable segmentation accuracy and generalization capabilities, making it possible to reconstruct training datasets based on target regions for semantic segmentation tasks. We leverage SAM-HQ\cite{samhq} to pre-segment UHR image masks and determine mask categories using image labels, thereby obtaining a set of region-category-labeled samples. This set serves as the foundation for region-based category resampling, which we refer to as \textbf{Semantic Reranking and Resampling method for Training Augmentation(SRR-TA)}.

As previously mentioned, the long-tail problem exacerbates the issues of high intraclass variance and low interclass variance. Inspired by \cite{protoseg}, which clusters features of the same category into multiple distinct subclusters during training, we extend this idea to enable the model to learn more fine-grained feature representations, thereby improving the accuracy of final feature classification. However, in \cite{protoseg}, prototypes are forcibly divided through clustering after feature extraction, leading to cluster centers that may lack clear semantic meaning. Additionally, using prototypes as classification criteria diminishes the strong guidance provided by labels. To address these issues, we propose the \textbf{Injection of General Representation Knowledge} module, which integrates textual information extracted from a remote sensing-specific vision-language model as an external knowledge source. This information is fused with the features extracted by the visual encoder, enhancing feature representation while preserving the original labels' influence on model learning.

In summary, our main contributions are as follows:
\begin{enumerate}
    \item We propose a novel Multi-Scale Anchored Region Sampling(MARS) method to supplant the conventional random cropping approach, enabling the anchor region to appear at any position within larger-scale cropped areas. This innovation provides the model training with a greater variety of angles and richer contextual information.
    \item We propose a training method SRR-TA that utilizes the SAM-HQ model to extract objects from satellite images and perform resampling, introducing category resampling in the UHR semantic segmentation field for the first time, which significantly enhancing the model's performance on long-tail distributed UHR satellite image datasets.
    \item For the first time, we have introduced the injection of general representation knowledge to enhance the accuracy of semantic segmentation of UHR images without text annotations for individual images. By integrating text related to remote sensing image categories and extracting text features using a domain-specific VLM, and fusing these features with visual features, we have demonstrated the efficacy of multi-modal feature fusion in the UHR domain.
\end{enumerate}

The remainder of this paper is organized as follows. In Section \ref{sec:relate_dwork}, we present an overview of the relevant work concerning UHR semantic segmentation and pre-trained visual-language models. Section \ref{sec:method} provides a detailed exposition of the methodology we propose. Section \ref{sec:exp} and Section \ref{sec:res} outlines the experiments we designed, along with a comparative analysis against other approaches. Finally, in Section \ref{sec:conclusion}, we summarize our method and offer a prospective outlook.

\section{Related Work}
\label{sec:relate_dwork}
\subsection{UHR Semantic Segmentation}
Numerous methods have been developed for semantic segmentation of UHR images. A common paradigm is to construct a multi-branch network, where each branch receives image inputs at different scales. After feature extraction, the features from different branches are fused to enhance segmentation results. GLNet\cite{GLNet} initially proposed a convolution-based dual-branch network that combines local and global features, while FctlNet\cite{Fctlnet} enhanced this approach by developing a triple-branch network to extract richer features. WiCoNet\cite{wconet} introduced a transformer-based dual-branch feature fusion network for the first time.

Building on the concept of multi-branch networks, GeoAgent\cite{GeoAgent} and PPN\cite{patchproposal} have designed additional modules to determine which image inputs to feed into the neural network, aiming to provide optimal information density and the most suitable receptive field size. GeoAgent\cite{GeoAgent} expanded on the dual-branch concept by integrating a reinforcement learning module to determine the optimal image size for the global branch. PPN \cite{patchproposal} introduced the Patch Proposal Network, which identifies regions that require refinement.

Some approaches are dedicated to processing entire ultra-high-resolution images in a single pass WSDNet\cite{wavelet} applied both Laplacian pyramid and wavelet transforms for feature extraction, using an image superpixel reconstruction task to assist segmentation. ISDNet\cite{isdnet} combined Laplacian pyramids of UHSR images with downsampled images, enhancing performance through hierarchical feature fusion and refinement. GPWFormer\cite{GPWFormer} used wavelet transforms to guide patch cropping grouping, followed by regrouping within clusters to reduce computational complexity.

\subsection{Pretrained Vision Models}
Pretrained visual models have achieved remarkable success in foundational visual tasks\cite{sam}\cite{CLIP}. Representative CLIP\cite{CLIP} has demonstrated exceptional image-text matching capabilities, offering features applicable to a diverse range of downstream tasks. In our previous work GeoRSCLIP\cite{GeoRSCLIP}, building on this, we integrated existing remote sensing datasets and added text descriptions to a large number of images. Using this extensive collection of image-text pairs, we trained a CLIP model specifically for the remote sensing domain. In addition to integrating satellite and aerial data, RemoteCLIP\cite{RemoteCLIP} incorporated drone images into the datasets to train the CLIP model. RS-CLIP\cite{rsclip} proposed a pseudo-label generation and curriculum learning fine-tuning strategy to transfer CLIP to the remote sensing domain.

Another notable example, SAM\cite{sam} and its derivative methods\cite{samhq}\cite{semantic_sam}\cite{presam}, has exhibited state-of-the-art performance in instance segmentation tasks for high-resolution images. SAM-HQ\cite{samhq} achieved impressive performance through a minimal adaptation that introduces a learnable High-Quality Output Token and a refined feature set fusion strategy. Semantic-SAM\cite{semantic_sam} is trained on multiple datasets with diverse semantic information and employs a multi-choice learning scheme to generate masks at multiple levels corresponding to various groundtruth masks. Currently, extensive efforts have been made to employ SAM in addressing challenges within the field of remote sensing \cite{sam_changedetect, sam_changeosm, sam_evaluating, mesam, sam4rs}. In this work, we utilize pretrained visual models to extract more robust features and perform high-quality image preprocessing.

\subsection{Long-Tail Problem}

The long-tail problem refers to the phenomenon in imbalanced datasets where a small number of head classes contain abundant samples, while the majority of tail classes have significantly fewer samples. Work\cite{lt-93} theoretically demonstrates that in a simple binary-class long-tail problem, the gradient update direction of one class is the opposite of that of the other class. Studies \cite{balancesample1, balancesample2, oversample, undersample}, investigate the impact of sampling strategies on model performance in long-tail classification tasks. Work\cite{lt-nc} proposes a center-collapse regularization method (CeCo) based on a fixed Equiangular Tight Frame (ETF) structure, which optimizes the distribution of feature centers to improve the recognition performance of tail classes. Work\cite{lt-ss} introduces a region-balancing strategy to address the long-tail problem in semantic segmentation. The study \cite{lt-or} proves that the positive samples of each class can be regarded as negative samples for other classes, leading to an inverse gradient effect for tail classes. By analyzing the characteristics of the long-tail problem in semantic segmentation tasks, we establish the theoretical foundation for our proposed method.

\section{PROPOSED METHOD}
\label{sec:method}
In this section, we initially present an overview of the primary workflow of our methodology in Section \ref{subsec:overview}. Subsequently, in Section \ref{subsec:MSAR}, we introduce the MSAR method that we have developed. Following this, in Section \ref{subsec:SRR-TA}, we provide a detailed exposition of the process of SRR-TA method. Finally, in Section \ref{subsec:inject}, we elaborate on the process of extracting text features using domain-specific VLMs for the injection of general representation knowledge.

\begin{figure*}[htbp]
    \centering
    \includegraphics[width=\textwidth]{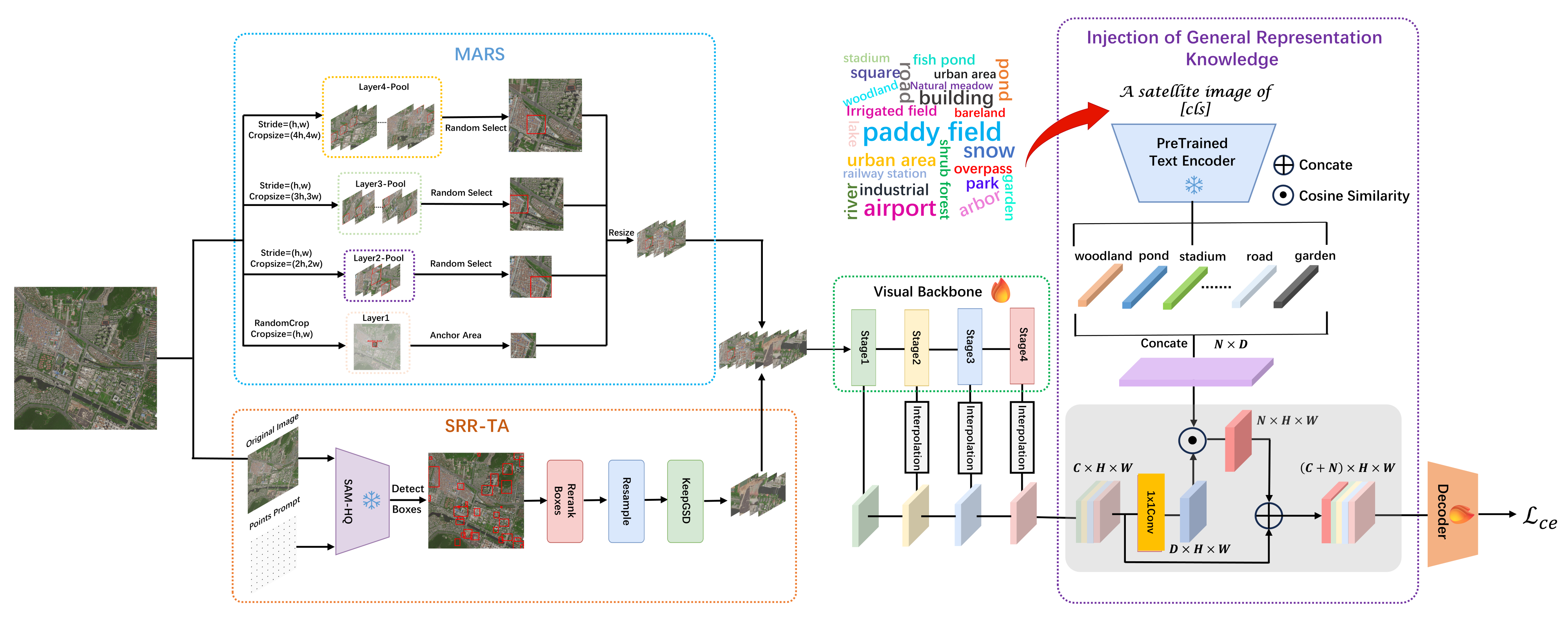} 
    \caption{The architectural overview of our proposed approach is presented, wherein MARS denotes Multi-Scale Anchored Region Sampling, and SRR-TA signifies Semantic Reranking and Resampling for Training Augmentation. }
    \label{fig:main}
\end{figure*}

\subsection{Overview of Our Method}
\label{subsec:overview}
Our method's training data for individual images comprises two components. One part is derived from the multi-scale cropping results of Multi-Scale Anchored Region Sampling(MSAR), and the other is obtained from Semantic Reranking and Resampling for Training Augmentation(SRR-TA), where detected bounding boxes are pre-extracted and undergo reordering and resampling. This process ensures that regions belonging to minority classes are more frequently sampled. We utilize the pre-trained GeoRSCLIP\cite{GeoRSCLIP} model to generate textual features, with its parameters frozen during training and not involved in optimization. During training, we simultaneously optimize the semantic segmentation network and the image-text feature fusion module.

We further integrate text descriptions of various remote sensing ground object categories. These descriptions are processed through a pre-trained text encoder to yield text features. These text features are then fused with visual features extracted from the training data by a visual backbone. The merged features are subsequently fed into a semantic segmentation decoder for pixel-level classification, culminating in the final segmentation outcome.

\subsection{Multi-Scale Anchored Region Sampling Method}
\label{subsec:MSAR}
Our proposed MSAR method effectively mitigates the impact of local specificity in large-scale ground objects and provides richer contextual information for each cropped region. Specifically, we proceed as follows:

Initially, a random region $z$ with shape $(h,w)$ is cropped from the original image, termed as the anchor area. Subsequently, based on the height and width of the anchor area, the original image is cropped at scales of 
$(k\times h,k\times w)$, where $k\in [2,3,4]$. This process retrieves cropped regions containing the anchor area at each scale, resulting in three distinct cropping pools. A region is randomly sampled from each pool, downscaled to the dimensions of the anchor area, and concatenated with the anchor area to form the training image. The process can be mathematically represented as Algorithm \ref{MSAR}.

Compared to random cropping, our method improves region access frequency and provides richer contextual information. Compared to multi-scale center cropping, where all other crops in a batch are centered around a specific region, our method selects random crops from the multi-scale cropping pool that include the anchor region when anchor region samples are similar. This approach provides a broader range of perspectives, enhancing region coverage while avoiding excessive repetition of training samples, which facilitates the learning of diverse features. The impact of MSAR on model learning can be observed in Table \ref{tab:ablation_study}. Neither random cropping nor multi-scale center cropping achieves the performance level of MSAR.

\begin{algorithm}
\caption{Multi-scale Anchored Region Sampling Algorithm}\label{MSAR}
\textbf{Input:} Original image \( I \) with height \( H \) and width \( W \), random anchor region size \( (h, w) \), scale factors \( k \in [2, 3, 4] \).

\textbf{1. Randomly select anchor area:} \\
\hspace*{0.5cm} \( z \leftarrow \text{random\_crop}(I, (h, w)) \)

\textbf{2. Generate multi-scale cropping pools:} \\
\hspace*{0.5cm} \textbf{for each} \( k \in [2, 3, 4] \) \textbf{do} \\
\hspace*{1.0cm} \textbf{$stride\_h, stride\_w \leftarrow h, w$} \\
\hspace*{1.0cm} \textbf{Initialize} \; $P_k \gets \emptyset$ \\
\hspace*{1.0cm} \textbf{for} \; $i = 0$ \; \textbf{to} \; $(H-kh)$ \; \textbf{with step} \; $stride\_h$ \\
\hspace*{1.5cm} \textbf{for} \; $j = 0$ \; to \; $(W-kw)$ \; with step \; $stride\_w$ \\
\hspace*{2.0cm} $if \; z \in I[i:i+kh, j:j+kw]$ \\
\hspace*{2.5cm} $P_k  \gets P_k \cup \{I[i:i+kh, j:j+kw]\}$ \\
\hspace*{1.5cm} \textbf{end for} \\
\hspace*{1.0cm} \textbf{end for} \\
\hspace*{0.5cm} \textbf{end for} 

\textbf{3. Randomly sample regions from each pool:} \\
\hspace*{0.5cm} \( \text{sampled\_regions} \leftarrow \{r_k \mid r_k \in P_k \text{ for } k \in [2, 3, 4]\} \)

\textbf{4. Downscale each sampled region to anchor size:} \\
\hspace*{0.5cm} \textbf{for each} \( r_k \in \text{sampled\_regions} \) \textbf{do} \\
\hspace*{1.0cm} \( r_k \leftarrow \text{resize}(r_k, (h, w)) \) \\
\hspace*{0.5cm} \textbf{end for}

\textbf{5. Concatenate anchor area with downscaled regions:} \\
\hspace*{0.5cm} \( \text{training\_image} \leftarrow \text{concat}(z, r_2, r_3, r_4) \)

\textbf{Output:} \( \text{training\_image} \)

\end{algorithm}

\subsection{Semantic Reranking and Resampling for Training Augmentation}
\label{subsec:SRR-TA}

\begin{figure}[htbp]
    \centering
    \includegraphics[width=\columnwidth]{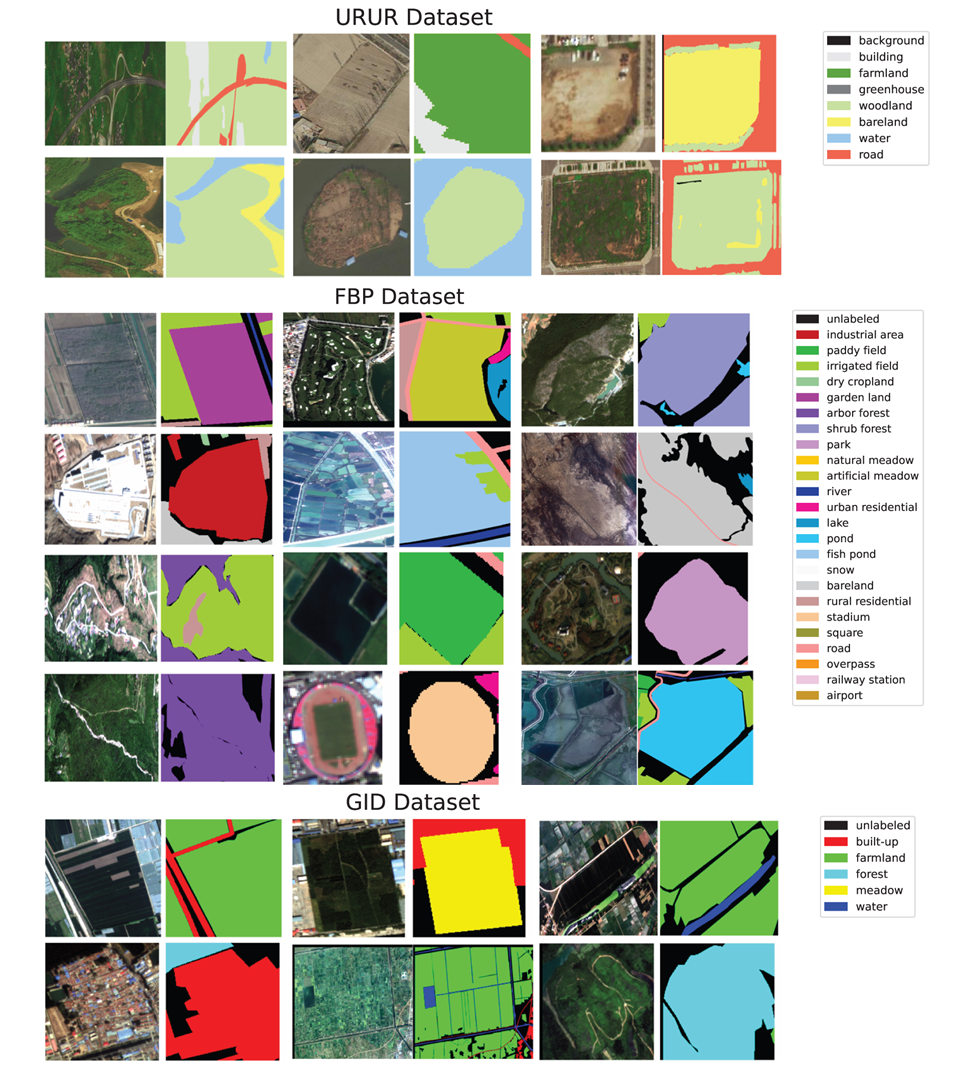} 
    \caption{Pre-extraction of multi-scale ground objects in three datasets.}
    \label{fig:crops}
\end{figure}

We first theoretically analyze the impact of the long-tail problem on gradient updates.
The Softmax Cross-Entropy loss function is defined by Equation \ref{eq:sft_ce} :

\begin{align}
    \mathcal{L} &= -\frac{1}{B} \sum_{n=1}^{B} \sum_{i=1}^{C} y_{n,i} \log p_{n,i}  \notag \\
    p_{x,i} &= \frac{e^{z_{x,i}}}{\sum_{j=1}^{C} e^{z_{x,j}}}
    \label{eq:sft_ce}
\end{align}

Where \( B \) is the batch size, \( C \) is the number of categories, \( y_{n,i} \) is the ground-truth one-hot label for the \( n \)-th pixel (or pixel block) in category \( i \), where \( y_{n,i} = 1 \) if the pixel belongs to class \( i \), otherwise \( y_{n,i} = 0 \). \( p_{n,i} \) is the probability of class \( i \) for the \( n \)-th pixel after Softmax normalization. 

The gradient computation is formulated as Equation \ref{eq:grad}.

\begin{align}
    \frac{\partial \mathcal{L}}{\partial z_{n,i}} &= \frac{1}{B} \sum_{n=1}^{B} (p_{n,i} - y_{n,i}) \notag \\
    &= \frac{1}{B} 
    \left( \sum_{n \in \mathcal{X}_i} (p_{n,i} - 1) + \sum_{n \notin \mathcal{X}_i} p_{n,i} \right)
    \label{eq:grad}
\end{align}

\( z_{n,i} \) is the logit (pre-Softmax output) for class \( i \) of the \( n \)-th pixel. \( \mathcal{X}_i \) represents the set of pixels belonging to class \( i \), i.e., \( \mathcal{X}_i = \{ n \mid y_{n,i} = 1 \} \). \( \frac{\partial \mathcal{L}}{\partial z_{n,i}} \) represents the gradient of the loss function with respect to \( z_{n,i} \).

From the above gradient computation formula, it can be observed that for any given category, the gradients generated by all pixels that do not belong to this category within a batch will have an opposing effect on the gradient update of this category. A balanced class distribution within the batch provides a regularization effect. If the class distribution within a batch is approximately balanced, the condition \ref{eq:lt_norm} holds. 

\begin{align}
    &\sum_{x \in \mathcal{X}} y_{x,k} \approx \sum_{x \in \mathcal{X}} y_{x,i}, \quad \forall i, k  \notag \\
    &\sum_{x \in \mathcal{X}} (p_{x,i} - y_{x,i}) \approx 0, \quad \forall i
    \label{eq:lt_norm}
\end{align}

Since \( p \) is the output of the Softmax function, its sum closely matches that of \( y \) under class-balanced conditions, providing a regularization effect. This ensures that gradient updates remain balanced across all class directions, contributing to more stable feature learning.

 In the UHR image setting, we extract multiple subimages from a single image to form a training batch. A portion of these sub-images is derived from MSAR to preserve the original class distribution of the image, while the remaining part comes from SAM-HQ\cite{samhq} detection results. Each image yields several object bounding boxes with primary class labels, and tail classes are explicitly selected as an additional part of the training data. We also explored a data augmentation method that specifically targets long-tail categories across the entire dataset. This approach maintains the cropped images from MSAR unchanged while selecting detected bounding boxes of tail categories from the SAM-HQ detections across the dataset as augmented data. A detailed comparison of both methods can be found in Section \ref{subsec:sota}. 

As illustrated in Figure \ref{fig:main}, we first generate prompts for each UHR image using points. 
We adopt a simple uniform distribution to generate prompt points. Specifically, we uniformly generate prompt points along both the horizontal and vertical coordinates, forming a grid of prompt points to serve as input for the SAM model. These prompt points are then fed into the model, where they pass through the prompt encoder, image encoder, and SAM decoder to produce masks for various scales of land covers. After obtaining the mask, the process involves first extracting the coordinates of all foreground pixels within the mask. Next, compute the minimum and maximum values of these coordinates to determine the bounding box. Specifically, identify the minimum x-coordinate ($x_{min}$), maximum x-coordinate ($x_{max}$), minimum y-coordinate ($y_{min}$), and maximum y-coordinate ($y_{max}$) of the foreground pixels. Finally, use these coordinates to generate a rectangular bounding box, where the top-left corner is located at $(x_{min}, y_{min})$ and the bottom-right corner is located at $(x_{max}, y_{max})$, thereby accurately representing the minimal bounding box that encompasses the entire mask.

Furthermore, we assign sampling priorities to each obtained region. For each region, we establish the following attributes: the primary category, which indicates which category predominantly constitutes the majority of the region; and category richness, which signifies the diversity of categories present within the region. Given that we have prior knowledge of the entire dataset's distribution, a region whose primary category is a minority class should be more likely to be selected during sampling. If two regions share the same primary category, the region with greater category richness should have a higher probability of being chosen. Thus, for each UHR remote sensing image, we obtain a set of training data that includes multiple complete ground objects with assigned sampling priorities. Consequently, we acquire multiple regions suitable for category resampling. The multi-scale ground object examples obtained are illustrated in Figure \ref{fig:crops}.

\subsection{Injection of General Representation Knowledge}
\label{subsec:inject}
Due to certain types of geographic objects frequently occupying major portions of UHR images, models struggle to learn features from categories that cover smaller overall areas, often leading to issues of class imbalance and confusion. Accordingly, we aim to extract text features for each category from a remote sensing-pretrained Vision and Language Model (VLM) as general representation knowledge. By fusing these text features with the visual representations extracted from the visual backbone, we obtain an integrated set of features. These features are then handed over to the decoder for pixel-level classification. The specific processing procedure is depicted in Figure \ref{fig:main}.

To obtain text features, we initially integrate common land cover categories found in satellite image. During this integration, we meticulously expanded upon certain broad categories, resulting in $K$ commonly recognized classes within satellite imagery. Adhering to the methodology of the CLIP\cite{CLIP} model, we employ the template "A satellite image of $\{\}$", where $\{\}$ is filled with the class name. The comprehensive list of these categories is presented in Table \ref{tab:categories}. For satellite image data with annotations limited to a few classes, we still utilize the text features of these $K$ categories as a general representation to fuse with visual representations. Based on our final integration, $K$ is determined to be 54.

\begin{table*}[hbp]
\centering
\caption{List of Categories}
\label{tab:categories}
\begin{tabular}{@{}ll@{}}
\toprule
\textbf{Category} & \textbf{Terms} \\
\midrule
Buildings & roof, building, built-up, construction, architecture, facility, house, skyscraper, rural residential, urban residential \\
\cmidrule{1-2}
Transportation & stadium, railway station, airport \\
\cmidrule{1-2}
Roads & street, road, highway, path, route, lane, avenue, way \\
\cmidrule{1-2}
Water Bodies & liquid, water, river, lake, pond, ocean \\
\cmidrule{1-2}
Barren & barren land, wasteland, unlabeled \\
\cmidrule{1-2}
Forest & woodland, jungle, bush, forest, woods, grove \\
\cmidrule{1-2}
Agriculture & farming, farmland, agrarian, ranching, agricultural land, irrigated field \\
\cmidrule{1-2}
Greenhouses & greenhouse, hothouse, glasshouse \\
\cmidrule{1-2}
Meadows & shrubs, meadow, herbs, grass, grassland, pasture, prairie, natural meadow, artificial meadow \\
\bottomrule
\end{tabular}
\end{table*}

After acquiring $K$ remote sensing geographic object descriptions, we put them into the text branch of GeoRSCLIP\cite{GeoRSCLIP} to get the text representation. Vision-language models typically aggregate features of an entire area into a single feature vector, whereas the backbone of a semantic segmentation model extracts pixel-wise feature maps. Let the visual representation extracted by the visual backbone be denoted as $\mathbb{R}_i^{h\times w\times c}$, and the text representation extracted by the VLM as $\mathbb{R}_t^{k\times d}$. We first align the visual features with the text features. Our experiments reveal that a simple single fully connected layer can achieve satisfactory results. Subsequently, we compute the cosine similarity between each pixel feature of the visual representation and each vector in $\mathbb{R}_t^{k\times d}$, thereby obtaining the fused features. We then concatenate the fused features with the visual representation to derive the final features used for decoding. This part of the operation can be represented by the following formula:

\begin{align}
    &\hat{R}_i^{h \times w \times d} = R_i^{h \times w \times c} \cdot W^{c\times d} \label{eq:align} \\
    &F_{i,j,k} = \frac{\hat{R}_i(i,j,:) \cdot R_t(k,:)}{\|\hat{R}_i(i,j,:)\| \cdot \|R_t(k,:)\|} , \quad F \in F^{h\times w \times k} \label{eq:fusion} \\
    &F_f = concate(F, R_i) , \quad F_f \in F_f^{h\times w \times (k+c)} \label{eq:concate}
\end{align}

Given the extracted feature $F_f$ from Equation (3), the final logits $Z$ are obtained using a $1 \times 1$ convolution:

\begin{equation}
    Z = F_f \ast W_{\text{logits}}, \quad Z \in \mathbb{R}^{h \times w \times N} \label{eq:Z}
\end{equation}

where $W_{\text{logits}} \in \mathbb{R}^{(k+c) \times N}$ is the weight of the $1 \times 1$ convolution layer, and $N$ represents the number of target classes.

The predicted probability for each class is computed using the softmax function:

\begin{equation}
    P(i,j,n) = \frac{\exp(Z(i,j,n))}{\sum_{m=1}^{N} \exp(Z(i,j,m))}, \quad P \in \mathbb{R}^{h \times w \times N} \label{eq:sotmax}
\end{equation}

The final loss function is the pixel-wise categorical cross-entropy loss:

\begin{equation}
    \mathcal{L} = - \sum_{i=1}^{h} \sum_{j=1}^{w} \sum_{n=1}^{N} Y(i,j,n) \log P(i,j,n) \label{eq:ce_loss}
\end{equation}

where $Y \in \{0,1\}^{h \times w \times N}$ is the one-hot ground truth segmentation mask.

\section{EXPERIMENTAL DATASETS AND SETUP}
\label{sec:exp}
In this section, we report the data and experimental setup utilized in our study. Initially, we present the datasets employed, which encompass three open-source datasets. Subsequently, we provide a detailed account of the experimental configuration and the metrics for evaluation.

\subsection{Dataset}

\begin{figure}[htbp] 
\centering
\includegraphics[width=\linewidth]{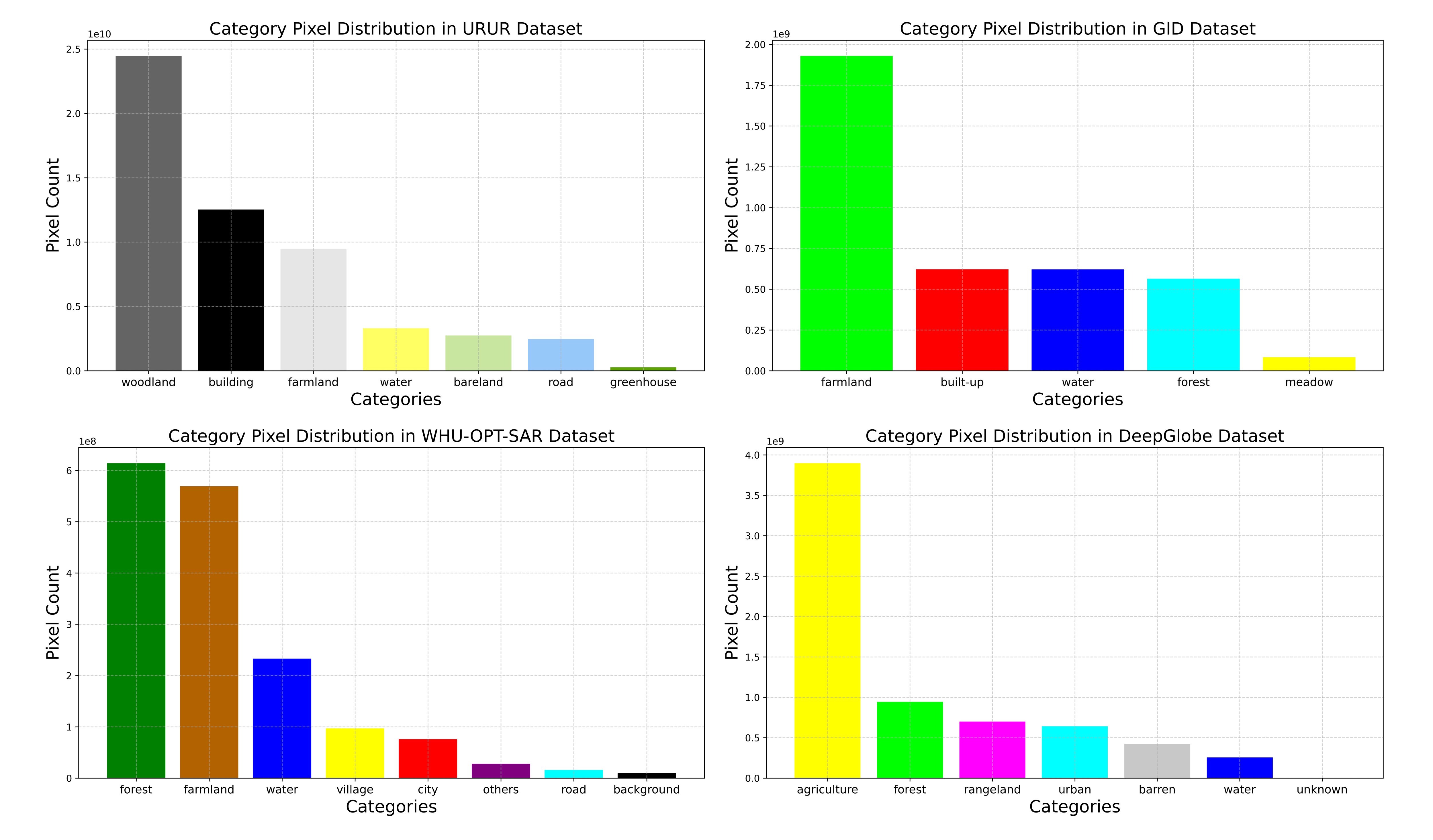} 
\caption{The distribution of class pixel counts in the training set of URUR, GID, WHU-OPT-SAR and DeepGlobe datasets.}
\label{fig:dist4}
\end{figure}

We selected five distinct datasets to substantiate the efficacy of our approach: a coarsely annotated dataset, GID\cite{GID}, utilizing its labeling of five land cover types as our ground truth; a high-resolution satellite image dataset with pixel-level fine annotations, URUR\cite{urur}, comprising a total of 3008 ultra-high-resolution satellite images; a dataset with a greater number of land cover category annotations, FBP\cite{FBP}, encompassing 24 foreground land cover types and one background class; a multi-mode high-resolution satellite image dataset, WHU-OPT-SAR\cite{whuopt}; a high-resolution rural land cover classification dataset DeepGlobe\cite{deepglobe}. The details of these datasets are presented as follows. The pixel-wise class distribution of these datasets is shown in Figures \ref{fig:pixel_dist} and \ref{fig:dist4}.

\subsubsection{URUR Dataset}

The URUR dataset \cite{urur} consists of 3,008 UHR satellite images (5120 × 5120 pixels) with pixel-level annotations. Collected from 63 cities, it features diverse resolutions and perspectives, making it the most contextually rich UHR dataset. The dataset contains seven foreground classes (building, farmland, greenhouse, woodland, bareland, water, road) and one background class. We adopt a 7:1:2 train-validation-test split.

\subsubsection{GID Dataset}

The GID dataset \cite{GID} includes 150 UHR satellite images (6800 × 7200 pixels) covering 50,000 km² in China, sourced from the Gaofen-2 satellite. It provides four spectral bands (NIR, RGB), but we use only RGB for our experiments. The dataset is divided into 120 training and 30 test images.

\subsubsection{FBP Dataset}

Building on GID, the FBP dataset \cite{FBP} expands the annotation set to 24 foreground land cover categories, achieving a 4-meter resolution over 50,000 km². It is currently the most finely classified UHR remote sensing dataset.

\subsubsection{WHU-OPT-SAR Dataset}
The WHU-OPT-SAR dataset \cite{whuopt} integrates optical (RGB, NIR) and SAR imagery, covering 50,000 km² in Hubei Province, China, at a 5-meter resolution. It contains 100 image pairs (5556 × 3704 pixels) with pixel-level annotations. We use only RGB channels for experiments and split the dataset into training (70\%), validation (10\%), and test (20\%).

\subsubsection{DeepGlobe Dataset}
The DeepGlobe dataset \cite{deepglobe} is a high-resolution rural land cover dataset, featuring 803 images (2448 × 2448 pixels, 50 cm resolution) covering 17,169 km². Since validation and test labels are not publicly available, we re-split the training set into training, validation, and test sets (7:1:2 ratio).

\subsection{Evaluation Metrics}

The mean Intersection over Union (mIoU) is a widely used metric in semantic segmentation to evaluate model performance by comparing the predicted segmentation map with the ground truth. It is derived from the confusion matrix and measures the overlap between the predicted and ground truth segmentations for each class.

For a given class $i$, the IoU is defined as the ratio of the intersection to the union of the predicted and ground truth pixels:

\begin{equation}
\begin{aligned}
\text{IoU}_i &= \frac{TP_i}{TP_i + FP_i + FN_i}
\end{aligned}
\label{eq:iou}
\end{equation}

where:
\begin{itemize}
    \item $TP_i$ (True Positives) is the number of pixels correctly classified as class $i$.
    \item $FP_i$ (False Positives) is the number of pixels incorrectly classified as class $i$, but which actually belong to a different class.
    \item $FN_i$ (False Negatives) is the number of pixels that belong to class $i$, but were incorrectly classified as another class.
\end{itemize}

In the context of the confusion matrix, for class $i$, the diagonal entry $M_{ii}$ represents the true positives ($TP_i$), the row entries $M_{ij}$ (for $j \neq i$) represent the false positives, and the column entries $M_{ki}$ (for $k \neq i$) represent the false negatives.

The mean IoU (mIoU) is computed by taking the average of the IoU values across all $C$ classes:

\begin{align}
\text{mIoU} &= \frac{1}{C} \sum_{i=1}^{C} \frac{M_{ii}}{M_{ii} + \sum_{j \neq i} M_{ij} + \sum_{k \neq i} M_{ki}}
\end{align}
\label{eq:miou}

where:
\begin{itemize}
    \item $M_{ii}$ is the number of true positives for class $i$.
    \item $M_{ij}$ represents the false positives for class $i$ (i.e., pixels of class $j$ predicted as class $i$).
    \item $M_{ki}$ represents the false negatives for class $i$ (i.e., pixels of class $i$ predicted as class $k$).
\end{itemize}

The mIoU is particularly advantageous as it provides a balanced evaluation of segmentation performance by accounting for both over-segmentation and under-segmentation errors across multiple classes.

\section{EXPERIMENTAL RESULTS}
\label{sec:res}

This section will present the experimental outcomes. We employ a sliding window prediction approach with a window size of $(512, 512)$ and a stride of $(341, 341)$ to forecast the entire UHR image. Initially, we compare our results with those of other State-of-the-Art (SOTA) methods on the datasets employed in Section \ref{subsec:sota}, encompassing both general semantic segmentation approaches and those specifically tailored for UHR remote sensing image segmentation. Subsequently, we conduct ablation studies to substantiate the efficacy of our methodology. Moving forward, we investigate the performance enhancement attributed to multi-scale random cropping and the utilization of SAM for resampling. Finally, we explore the facilitative role of features from pre-trained vision-language models(VLMs) on the UHR semantic segmentation task from various perspectives. 

\subsection{Comparison with State-of-the-Art (SOTA) Methods}
\label{subsec:sota}
We initially compared the performance of our proposed SRMF with that of general semantic segmentation networks. To ensure the completeness of our experiments, we selected CNN-based architectures such as FPN\cite{fpn}, UNet\cite{unet}, PSPNet\cite{pspnet}, and Deeplabv3+\cite{deeplabv3p}, as well as Transformer-based models such as Segformer\cite{segformer} and Swin Transformer\cite{swintrans}. Given that Segformer is introduced as foundational models, we adhered to the practices within the MMSEG\cite{mmseg} library during the experimental process, using the semantic segmentation head of UperNet\cite{upernet} as the decoder and augmenting it with an FCN\cite{fcn} head for auxiliary training. The implementation of FPN\footnotemark[1] followed \cite{fpn}, while other models were implemented using the approaches provided within MMSEG\cite{mmseg}. The experimental setup was in accordance with Section \ref{sec:res}.

\begin{figure*}[htbp]
    \centering
    \includegraphics[width=\textwidth]{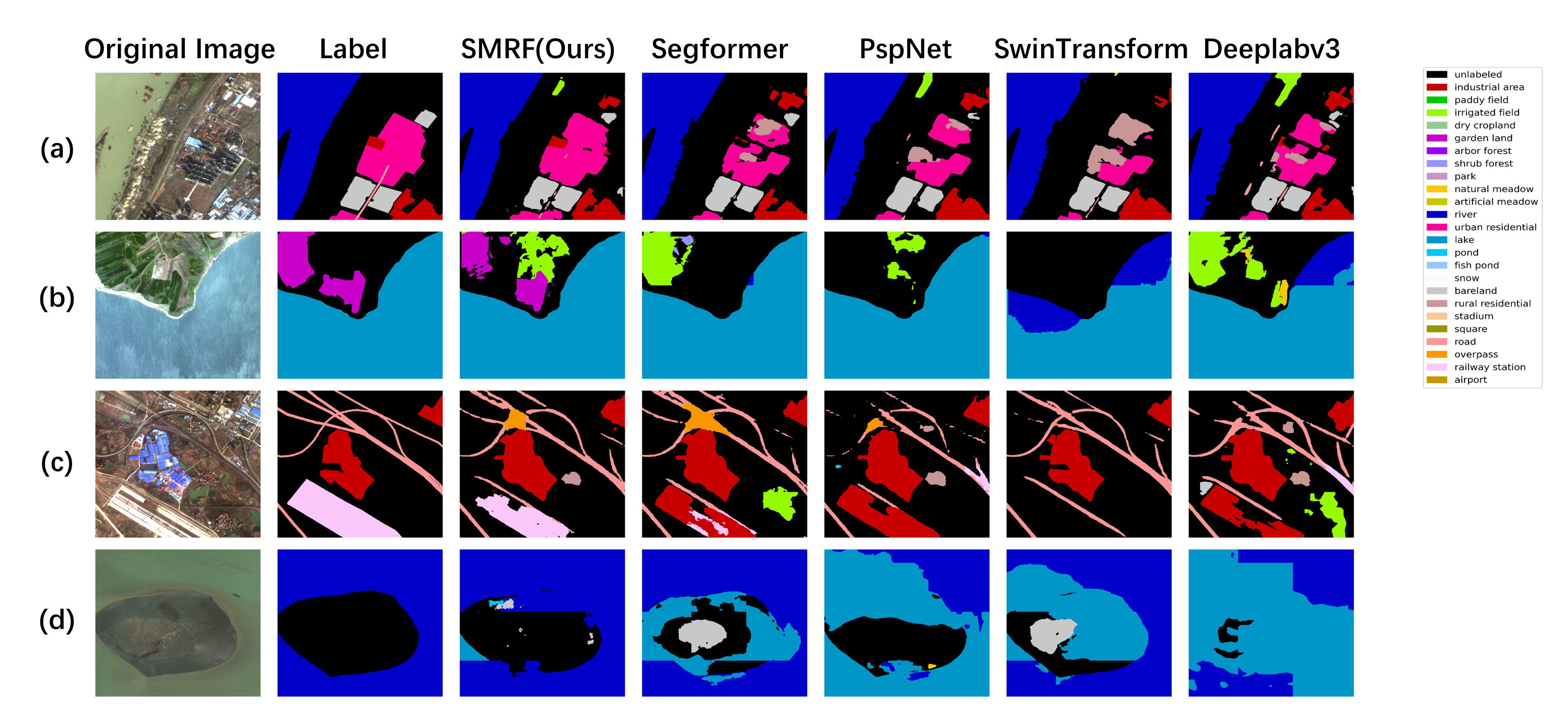} 
    \caption{A comparative analysis of the SRMF method with other state-of-the-art (SOTA) approaches on the FBP dataset is presented. In (a), it can be observed that our method performs better on the two easily confused classes of urban residential and rural residential, with fewer misclassified areas. In (b), despite the presence of misclassifications, our model demonstrates a clear advantage over other models in distinguishing between arbor forest and irrigated field classes. In (c), while other models confuse railway station and industrial area, our model accurately differentiates between them. In (d), other models confuse the classes of river and pond, but our model does not require a broader context to accurately distinguish the features of river and pond, showcasing the superiority of our model in differentiating classes with very similar characteristics.
}
    \label{fig:fbp_final}
\end{figure*}

\begin{figure*}[htbp]
    \centering
    \includegraphics[width=\textwidth]{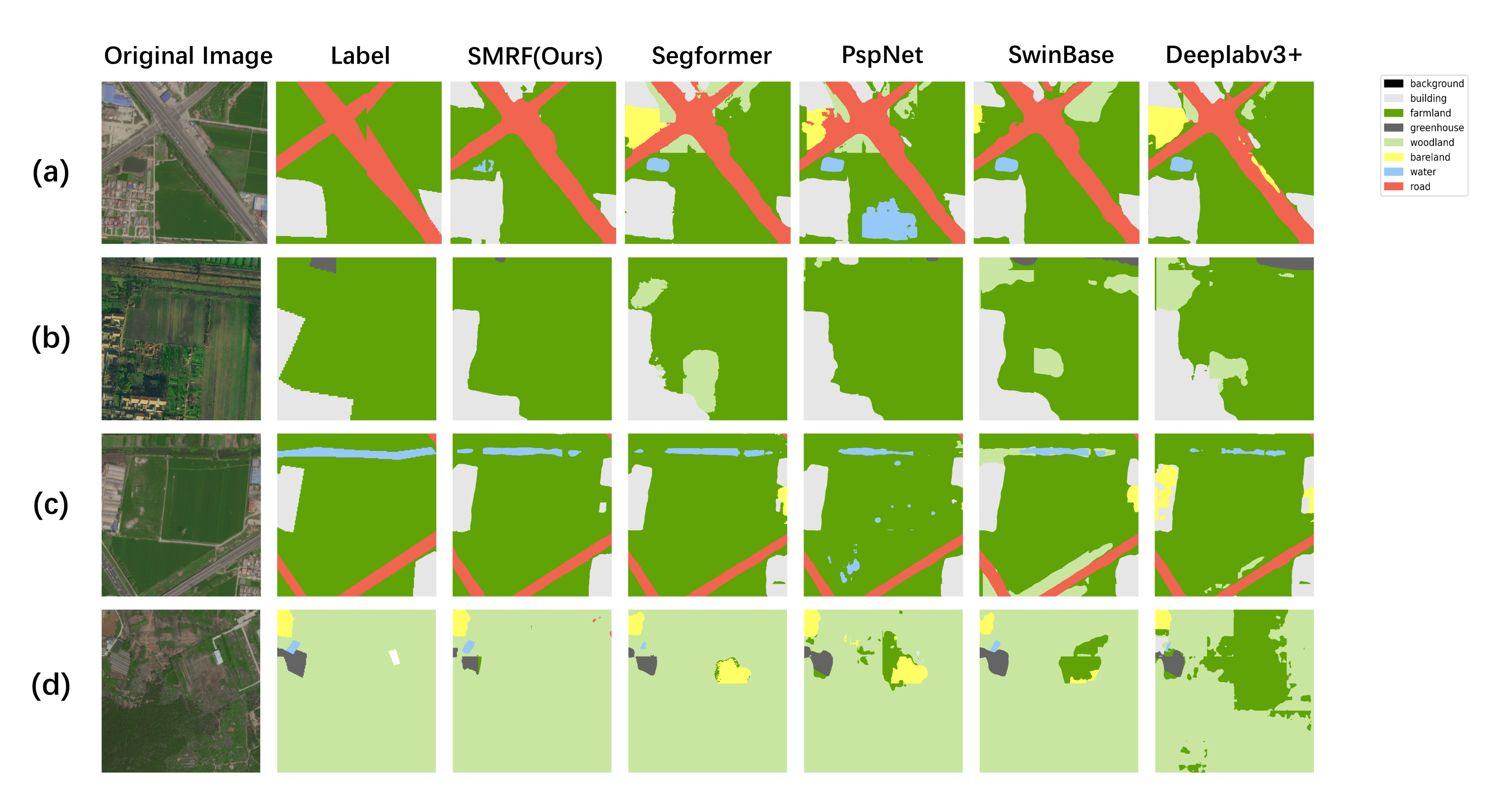} 
    \caption{A comparative assessment of the SRMF method against other state-of-the-art (SOTA) methods on the URUR dataset is depicted. From (a), (b), and (d), it is observable that other approaches exhibit confusion between the classes of woodland and farmland, whereas our method effectively discriminates the characteristics of these two land cover types. Furthermore, observations from (a), (c), and (d) reveal that barren land is readily confounded with buildings by other methods. In contrast, our approach successfully distinguishes the features of these two categories.
}
    \label{fig:urur_final}
\end{figure*}

Subsequently, we will juxtapose our methodology with those tailored for the semantic segmentation of UHR satellite images. The comparative analysis encompasses the following approaches: the dual-path network with enhanced refinement in two phases GLNet\cite{GLNet}, the contextual feature integration network WiCoNet\cite{wconet} leveraging the transformer architecture, the FtclNet\cite{Fctlnet} incorporating multi-scale contextual feature fusion, and the ISDNet\cite{isdnet} that employs a combination of shallow and deep networks along with relation-aware feature fusion. The detailed experimental results are presented in Table~\ref{main_results}. The performance of the SMRF method on the FBP and URUR datasets is illustrated in Figures \ref{fig:fbp_final} and \ref{fig:urur_final}, respectively. It can be observed that our approach significantly reduces misclassifications in relatively minority classes, such as the barren land category in the URUR dataset and the pond and railway station categories in the FBP dataset, thereby achieving improved segmentation outcomes for these underrepresented classes.

Finally, we investigate the long-tail data augmentation method at the dataset level proposed in Section \ref{subsec:SRR-TA}, referred to as SRMF-LT. As observed, SRMF-LT significantly underperforms the batch-internal class balancing method across the URUR, GID, and FBP datasets. This is because within a training batch, the globally defined tail categories may not necessarily align with the tail categories present in that specific batch. Consequently, this increases intrabatch class imbalance, leading to unstable training, hindered feature learning, and a higher likelihood of feature collapse. Specifically, certain class logits may become excessively close to 0, resulting in numerical overflow when computing the loss function. This overflow then propagates to the computation of gradients, ultimately causing numerical instability and training failure.

\begin{table*}[!t]
\renewcommand{\arraystretch}{1.3}
\caption{The semantic segmentation results across five datasets.}
\label{main_results}
\centering
\begin{tabular}{|c|c|c|c|c|c|c|c|c|}
\hline
\multirow{2}{*}{Method} & \multirow{2}{*}{Backbone} & \multirow{2}{*}{Parameters(M)} & \multirow{2}{*}{FLOPs(G)} & \multicolumn{1}{c|}{URUR} & \multicolumn{1}{c|}{GID} & \multicolumn{1}{c|}{FBP} & \multicolumn{1}{c|}{WHU-OPT-SAR} & \multicolumn{1}{c|}{DeepGlobe} \\
\cline{5-9} 
 & & & & mIoU (\%) & mIoU (\%) & mIoU (\%) & mIoU (\%) & mIoU (\%) \\
\hline
\multicolumn{9}{|c|}{General-purpose semantic segmentation methods} \\
\hline
UNet\cite{unet} & UNet-s5-d16 & 28.99 & 203.00 & 42.71 & 74.46 & 44.19 & 53.33 & 70.33\\
PSPNet\cite{pspnet} & Resnet-101 & 65.60 & 256.00 & 43.40 & 75.89 & 51.09 & 55.08 &73.38\\
FPN\cite{fpn} & Resnet-101 & 46.14 & 163.85 & 39.70 & 56.48 & 20.612 & 49.11 & 55.92\\
DeepLabv3+\cite{deeplabv3p} & Resnet-101 & 60.21 & 254.00 & 42.34 & 76.04 & 50.92 & 55.51 & 73.29\\
SwinTransformer\cite{swintrans} & Swin-Base & 120.00 & 298.00 & 44.31 & 76.19 & 50.63 & 55.96 & \textbf{73.83} \\
Segformer\cite{segformer} & MiT-B5 & 81.98 & 74.58 & 43.69 & 76.39 & 50.70 & \textbf{56.09} & 73.55\\
\hline
\multicolumn{9}{|c|}{UHR satellite image-specific semantic segmentation methods} \\
\hline
GLNet\cite{GLNet} & FPN-ResNet-50 & 37.31 & 140.26 & 41.20 & 65.14 & 32.05 & 42.24 & 69.18 \\
WiCoNet\cite{wconet} & FCN-ResNet-50 & 38.25 & 41.87 & 40.34 & 75.69 & 50.54 & 47.16 & 69.54\\
FCtLNet\cite{Fctlnet} & Segformer-MIT-B5 & 155.33 & 660.00 & 43.10 & 72.18 & 43.12 & 45.73 & 70.53\\
ISDNet\cite{isdnet} & ResNet50 & 17.78 & 27.55 & 45.80 & 75.60 & 49.32 & 50.33 & 70.69 \\
BEDSN\cite{bedsn} & Custom Backbone & 63.85 & 64.23 & 44.15 & 69.86 & 34.47 & 49.04 & 57.46 \\
WSDNet\cite{urur} & Deeplabv3plus-ResNet-18 & - & - & 46.90 & - & - & - & - \\
\hline
\multicolumn{9}{|c|}{The SRMF Approach Based on Data Augmentation and Multimodal Fusion} \\
\hline
SRMF & MiT-B5 & 83.04 & 92.93 & \textbf{47.02} & \textbf{77.47} & \textbf{51.68} & 55.31 & 71.06\\
SRMF-LT & MiT-B5 & 83.04 & 92.93 & 44.99 & 76.75 & 48.30 & 55.58 & 71.50\\
\hline
\end{tabular}
\end{table*}

\footnotetext[1]{https://github.com/jwyang/fpn.pytorch.git}

\subsection{Investigation of MSAR and SRR-TA}
\label{subsec:MSARandResample}
In this section, we investigate the impact of training data combination methods to ascertain the critical factors for semantic segmentation of UHR satellite imagery. We employed three distinct data combination strategies:

\subsubsection{MSAR \& GSD-preserving}
\label{subsec:exp1}
As delineated in Section A, we initially applied our proposed MSAR method to the original images to generate a set of images that encapsulate contextual information. In this experiment, the cropping scales were set to $[1, 2, 3, 4]$, yielding a dataset referred to as training data group $b_{mc}$. Subsequently, from the data pre-extracted using the SAM-HQ method, for each original image, we selectively sampled from the minority classes. If the area's dimensions exceeded the training size, we maintained the original ground sampling distance(gsd) and randomly cropped images of the training size from this area to form part of the training data. Conversely, if the area's dimensions were smaller than the training size, we commence the cropping process from the top-left corner of the cropping area, extracting an image of the same dimensions as the training size to serve as a training image, resulting in training data group $b_{gp}$, which was then combined with group $b_{mc}$ to form a composite training batch.

\subsubsection{MSAR \& w/o GSD-preserving}
\label{subsec:exp2} 
Building upon the foundation of Experiment MSAR\& GSD-preserving, the multi-scale cropping step remained unchanged. However, when utilizing the data pre-extracted with the SAM-HQ method, the gsd was no longer preserved. Instead, the target areas were directly resized to the training dimensions to obtain $b_{ngp}$. $b_{mc}$ and $b_{ngp}$ were then concatenated to form a composite training batch.

\subsubsection{WG-ResCro}
\label{subsec:exp3}
This experiment eschewed the MSAR without fixed areas, opting to utilize the data pre-extracted by SAM-HQ in its entirety. This method, referred to as weighted gsd-preserving resizing and cropping for training data(WG-ResCro), is specifically described as follows. In this experiment, we employed an exponentially weighted resampling method to generate the training data. This method performs weighted random sampling from an array \(\mathbf{a})\) of length \( n \), where the probability of selecting each element is inversely proportional to its index in the array. The procedure can be described as Algorithm~\ref{experiment3}.

The experimental results for the GID, URUR, and FBP datasets are presented in Table~\ref{tab:datasample_gid}, \ref{tab:datasample_urur}, and \ref{tab:datasample_fbp}, respectively. From these sets of experimental outcomes, it is evident that maintaining the ground sampling distance is crucial during the process of category selection for resampling the training data. Through comparative experiments w/o GSD-preserving, we identified several factors influencing the impact of SAM-HQ-detected regions on model performance. The primary factor is the number of long-tail category regions detected by SAM-HQ\cite{samhq}. Since we did not observe a specific spatial distribution pattern for long-tail categories, the detected regions predominantly belong to head categories. Our approach primarily adjusts the priority of selecting detected long-tail regions for training. This phenomenon is particularly evident in the \textbf{park, square, and snow} categories of the FBP dataset. These categories exhibit strong performance under the \textbf{with GSD-preserving} setting but experience a sharp decline in the \textbf{without GSD-preserving} setting. The reason for this degradation is that the number of SAM-HQ detected regions for \textbf{park, square, and snow} is only 16, 10, and 0, respectively. Regardless of how sampling is balanced, such a small number of training samples is far from sufficient for effective model learning. The GSD-preserving operation allows training samples to be drawn from a broader spatial range, thereby increasing the likelihood of sampling long-tail categories.

Another critical factor is that our GSD-preserving method effectively leverages the local semantic continuity of remote sensing satellite images. For instance, \textbf{farmland} and \textbf{woodland} are often located near water sources, while \textbf{park, square, and station} are typically surrounded by \textbf{urban residential} areas. Our experiments validated this theory. For example, we observed strong relationships between \textbf{farmland and water} in the GID dataset table\ref{tab:datasample_gid}, \textbf{woodland and water} in the URUR dataset table\ref{tab:datasample_urur}, and \textbf{station, rural residential, road, overpass, and urban residential} in the FBP dataset table\ref{tab:datasample_fbp}. These categories are not only long-tail classes but also exhibit strong local semantic continuity. As a result, although absolute balance was not achieved in the training data, these categories experienced simultaneous improvements during the learning process.

Contrasting the results from Experiment \ref{subsec:exp1} and Experiment \ref{subsec:exp3} reveals that the WG-ResCro method indeed outperforms MSAR \& GSD-preserving on long-tail categories such as road, barren land, and greenhouse. However, it exhibits a significant performance decline on head categories like farmland and woodland. This is because head categories typically consist of large land-cover objects, and their segmentation heavily relies on the availability of sufficient training samples. The over-sampling of long-tail categories in WG-ResCro resulted in an insufficient number of samples for head categories. We believe that with a more balanced training sample distribution, further performance improvements could be achieved. As for the FBP dataset, the presence of multiple small-sized land-cover objects among the long-tail categories makes the influencing factors more complex. Compared to MSAR \& GSD-preserving, the WG-ResCro method did not achieve performance improvements for park, snow, and railway station, primarily due to the limited number of regions detected by SAM-HQ. However, for categories with distinct features such as airport, stadium, and fish pond, WG-ResCro achieved substantial performance gains. These categories do not rely heavily on extensive contextual information, allowing them to extract features effectively even without it. Consequently, after eliminating the pixel loss caused by MSAR-induced rescaling, their performance improved.

Overall, the impact of SAM-HQ-detected regions on model performance is complex. When sampling to address the long-tail problem, factors such as object size, dependency on contextual information, and local semantic relationships between categories must be considered. The current GSD-preserving method achieves the best overall performance on average; however, other sampling strategies still demonstrate advantages and hold potential for further enhancement.

\begin{table*}[htbp]
\centering
\caption{The results of the three data sampling methods on the GID dataset.}
\label{tab:datasample_gid}
\begin{tabular}{ccccc ccc}
\toprule
\multicolumn{2}{c}{\textbf{Experiments}} & \multicolumn{5}{c}{\textbf{Per-class iou (\%)}} & \multicolumn{1}{c}{\textbf{Metrics}} \\
\cmidrule(lr){1-2}\cmidrule(lr){3-7} \cmidrule(lr){8-8}
\textbf{Method} & \textbf{Backbone} & \textbf{Built-up} & \textbf{Forest} & \textbf{Farmland} & \textbf{Meadow} & \textbf{Water} & \textbf{mIoU} \\
\midrule
\multicolumn{8}{c}{\textbf{Investigation of MSAR and GSD-preserving Method}} \\
\midrule
MSAR \& GSD-preserving & MIT-B5 & 79.46 & 81.55 & 77.86 & 54.70 & 91.43 & 77.00 \\
MSAR \& w/o GSD-preserving & MIT-B5 & 79.75 & 81.48 & 77.16 & 55.17 & 91.26 & 76.96 \\
WG-ResCro & MIT-B5 & 79.60 & 81.69 & 77.40 & 55.53 & 91.30 & 77.11 \\
\midrule
\multicolumn{8}{c}{\textbf{The Impact of Pre-Trained VLMs Features}} \\
\midrule
Txt Feature & MIT-B5 & 79.91 & 81.51 & 77.70 & 56.66 & 91.55 & 77.47 \\
Img Feature & MIT-B5 & 79.62 & 81.30 & 77.31 & 54.54 & 90.75 & 76.70 \\
Txt-Img Feature & MIT-B5 & 79.74 & 81.41 & 77.67 & 55.00 & 90.95 & 76.95 \\
\bottomrule
\end{tabular}
\end{table*}

\setlength{\tabcolsep}{4pt}
\begin{table*}[tbp]
\centering
\caption{The results of the three data sampling methods on the URUR dataset.}
\label{tab:datasample_urur}
\begin{tabular}{ccccc ccccc c}
\toprule
\multicolumn{2}{c}{\textbf{Experiments}} & \multicolumn{8}{c}{\textbf{Per-class iou (\%)}} & \multicolumn{1}{c}{\textbf{Metrics}} \\
\cmidrule(lr){1-2}\cmidrule(lr){3-10} \cmidrule(lr){11-11}
\textbf{Method} & \textbf{Backbone} & \textbf{Background} & \textbf{Building} & \textbf{Farmland} & \textbf{Grennhouse} & \textbf{Woodland} & \textbf{Bareland} & \textbf{Water} & \textbf{Road} & \textbf{mIoU} \\
\midrule
\multicolumn{11}{c}{\textbf{Investigation of MLMC Cropping and GSD-preserving Method}} \\
\midrule
MSAR \& GSD-preserving & MIT-B5 & 1.14 & 72.78 & 75.76 & 43.58 & 48.09 & 32.28 & 49.26 & 49.84 & 46.59 \\
MSAR \& w/o GSD-preserving & MIT-B5 & 0.01 & 71.78 & 73.28 & 43.73 & 45.53 & 36.52 & 43.70 & 50.21 & 45.59 \\
WG-ResCro & MIT-B5 & 0.00 & 72.67 & 71.13 & 44.80 & 43.77 & 35.80 & 44.85 & 50.01 & 45.38 \\
\midrule
\multicolumn{11}{c}{\textbf{The Impact of Pre-Trained VLMs Features}} \\
\midrule
Txt Feature & MIT-B5 & 1.06 & 72.40 & 75.91 & 45.19 & 47.77 & 33.15 & 51.51 & 49.18 & 47.02 \\
Img Feature & MIT-B5 & 0.73 & 72.03 & 74.65 & 44.31 & 46.58 & 35.97 & 49.22 & 48.17 & 46.46 \\
Txt-Img Feature & MIT-B5 & 1.34 & 71.83 & 71.48 & 44.68 & 44.85 & 36.01 & 42.90 & 49.59 & 45.34 \\
\bottomrule
\end{tabular}
\end{table*}

\subsection{Ablation Study}

In this section, we conduct experiments on each of the individual modules proposed to validate their effectiveness. 

\subsubsection{Configuration Details}

For the multi-layer multi-scale random cropping method we introduced, we perform cropping on the anchor area of each image at scales expanded by factors of 2, 3, and 4, and concatenate these to the anchor area to obtain a set of training data encompassing four scales. The ablation study on the scale factors is presented in Table \ref{tab:scale_factor}. Regarding the category resampling approach we proposed, for each UHR image, we select the top $K$ detected land cover bounding boxes according to their ranking. For the GID and URUR datasets, $K$ is set to 4; for the FBP dataset, $K$ is set to 8.

\begin{algorithm}
\caption{WG-ResCro}\label{experiment3}
\textbf{Input:} array \( arr \)(has been sorted in descending order of sampling priority.
), number of samples \( x \), array length factor \( arr\_len\_factor = 0.07 \)

\textbf{1. Calculate the length of the array:} \\
\hspace*{0.5cm} \( arr\_len \leftarrow \text{len}(arr) \)

\textbf{2. Compute the constant:} \\
\hspace*{0.5cm} \( C \leftarrow arr\_len \times arr\_len\_factor \)

\textbf{3. Assign weights to each element:} \\
\hspace*{0.5cm} \( w_i \leftarrow \frac{1}{i + C} \), where \( i = 0, 1, 2, \dots, arr\_len - 1 \)

\textbf{4. Normalize the weights to create a probability distribution:} \\
\hspace*{0.5cm} \( p_i \leftarrow \frac{w_i}{\sum_{j=0}^{arr\_len - 1} w_j} \)

\textbf{5. Sample \( x \) crops from the array based on the probability distribution:} \\
\hspace*{0.5cm} \( sampled\_crops \leftarrow \text{np.random.choice}(arr, size=x, p=p_i, replace=True) \)

\textbf{6. Initialize an empty training set:} \\
\hspace*{0.5cm} \( \text{train\_set} \leftarrow \text{[]} \)

\textbf{7. For each crop in selected crops:} \\
\hspace*{0.5cm} \textbf{for each} \( \text{crop} \in \text{selected\_crops} \) \textbf{do} \\
\hspace*{1.0cm} \( \text{crop1} \leftarrow \text{resize}(\text{crop}, \text{train\_size}) \) \\
\hspace*{1.0cm} \( \text{crop2} \leftarrow \text{random\_crop}(\text{crop}, \text{train\_size}) \) \\
\hspace*{1.0cm} \( \text{train\_set.expand}(\textbf{[}\text{crop1}, \text{crop2}\textbf{]}) \) \\
\hspace*{0.5cm} \textbf{end for}

\textbf{Output:} \( train\_set \)

\end{algorithm}

\begin{table*}[t]
    \centering
    \caption{Performance comparison under different scale factors across datasets.}
    \label{tab:scale_factor}
    \begin{tabular}{c|cccc}
        \toprule
        Dataset \textbackslash\ Scale Factors & 1 & 1,2 & 1,2,3 & 1,2,3,4 \\
        \midrule
        URUR & 44.58 & 45.56 & 45.23 & \textbf{47.02} \\
        DeepGlobe & 70.08 & 70.32 & 70.04 & \textbf{71.06} \\
        \bottomrule
    \end{tabular}
\end{table*}

\begin{table*}[htbp]
\centering
\caption{Quantitative results of the ablation study on the considered datasets.}
\label{tab:ablation_study}
\begin{threeparttable}
\begin{tabular}{lcccccccc}
\toprule
\textbf{Dataset} & \textbf{Method} & \multicolumn{4}{c}{\textbf{Components}} & \textbf{mIoU (\%)} \\
\cmidrule(lr){3-6}
 & & \textbf{MARS} & \textbf{CenMC} & \textbf{GSD-preserving} & \textbf{TxT-Feature} & \\
\midrule
\multirow{6}{*}{\textbf{GID}} & \multirow{6}{*}{Segformer-MIT-B5}
& \xmark & \xmark & \xmark & \xmark & 76.81 \\
& & \xmark & \cmark & \xmark & \xmark & 76.44 \\
& & \cmark & \xmark & \xmark & \xmark & 76.65 \\
& & \cmark & \xmark & \cmark & \xmark & 77.00 \\
& & \cmark & \xmark & \xmark & \cmark & 76.76 \\
& & \cmark & \xmark & \cmark & \cmark & \textbf{77.47} \\
\midrule
\multirow{6}{*}{\textbf{URUR}} & \multirow{6}{*}{Segformer-MIT-B5}
& \xmark & \xmark & \xmark & \xmark & 43.69 \\
& & \xmark & \cmark & \xmark & \xmark & 45.64 \\
& & \cmark & \xmark & \xmark & \xmark & 45.86 \\
& & \cmark & \xmark & \cmark & \xmark & 46.59 \\
& & \cmark & \xmark & \xmark & \cmark & 45.29 \\
& & \cmark & \xmark & \cmark & \cmark & \textbf{47.02} \\
\midrule
\multirow{6}{*}{\textbf{FBP}} & \multirow{6}{*}{Segformer-MIT-B5}
& \xmark & \xmark & \xmark & \xmark & 50.70 \\
& & \xmark & \cmark & \xmark & \xmark & 50.06 \\
& & \cmark & \xmark & \xmark & \xmark & 51.21 \\
& & \cmark & \xmark & \cmark & \xmark & 51.27 \\
& & \cmark & \xmark & \xmark & \cmark & 51.00 \\
& & \cmark & \xmark & \cmark & \cmark & \textbf{51.68} \\
\midrule
\multirow{3}{*}{\textbf{WHU-OPT-SAR}} & \multirow{3}{*}{Segformer-MIT-B5}
& \xmark & \xmark & \xmark & \xmark & 56.09 \\
& & \cmark & \xmark & \cmark & \xmark & \textbf{56.10} \\
& & \cmark & \xmark & \cmark & \cmark & 55.31 \\
\midrule
\multirow{3}{*}{\textbf{DeepGlobe}} & \multirow{3}{*}{Segformer-MIT-B5}
& \xmark & \xmark & \xmark & \xmark & \textbf{73.55} \\
& & \cmark & \xmark & \cmark & \xmark & 71.75 \\
& & \cmark & \xmark & \cmark & \cmark & 71.06 \\
\bottomrule
\end{tabular}
\begin{tablenotes}
    \item \textbf{Note:} The table displays quantitative results for different datasets with ablation study of MSAR-Crop, Resample, and TxT-Feature components. The CenMC means Centric Multi-Crop.
\end{tablenotes}
\end{threeparttable}
\end{table*}

\subsubsection{Quantitative Analysis}

The results of the ablation study are presented in Table \ref{tab:ablation_study}, where our method has achieved performance improvements across three datasets. Although the MSAR method led to a decrease in performance on the GID dataset, it achieved a performance enhancement of $2.17$ on the URUR dataset, with all categories showing improved segmentation outcomes. Furthermore, it realized a performance improvement of $0.51$ on the FBP dataset.

\subsubsection{Analysis of the MSAR Approach's Effectiveness}

Our cropping method, which merely expands the ground area covered by the input data without any feature fusion between sampled images, has led to performance enhancement. This indicates that the ground area covered by training images is also a significant factor affecting the task of UHR satellite image segmentation. A larger ground area, under fixed-size model training, results in training labels being downsampled and accuracy lost. However, it simultaneously provides richer land cover information and category details on a single training instance, thereby enhancing segmentation outcomes to a certain extent. As reflected in the experimental results, the GID dataset is primarily a coarsely annotated dataset, with training data sourced exclusively from the Gaofen-2 satellite, sampling only 60 cities in China, and thus offering relatively low feature complexity, where a larger ground area provides no significant benefit. In contrast, the URUR dataset encompasses a greater variety of satellite images and broader regions, with more complex land features. Providing a larger ground area for training in this case significantly enhances the model's ability to learn diverse features, aiding in the acquisition of more robust characteristics.

\subsubsection{Rich and Solid Pre-trained Knowledge from Text Features}

Our text features provide a very rich and solid foundation of pre-trained knowledge information. As shown in the \textbf{The Impact of Pre-Trained VLMs Features} section of Tables \ref{tab:datasample_gid}, \ref{tab:datasample_fbp}, \ref{tab:datasample_urur}, we demonstrate the effectiveness of using textual features as external knowledge. Furthermore, our ablation study \ref{tab:ablation_study} confirms that performance improvement can be achieved without the need for complex multimodal feature alignment modules, as a simple fully connected layer for feature mapping is sufficient. Since the training data for the pre-trained model offers different perspectives of land cover, the pre-training data contains a richer variety of land cover samples, with regional information from around the world infused into the pre-trained model's text features. As we inject text information into every location on the feature map, this information can be accurately integrated into every pixel of the image, greatly improving the segmentation outcomes.

However, we found that textual features became ineffective on the WHU-OPT-SAR\cite{whuopt} and DeepGlobe\cite{deepglobe} datasets. This is primarily because the integrated textual descriptions did not cover all categories present in these datasets, preventing the model from learning appropriate class centers. The categories such as others, rangeland, and village were not incorporated into our textual descriptions, leading to a decline in performance. This observation is corroborated by the ablation experiments presented in Table \ref{tab:ablation_study} for these two datasets, where the removal of text feature augmentation resulted in performance improvement.

Our ablation experiments further demonstrate that the effectiveness of textual features depends on data augmentation from SAM-HQ\cite{samhq} detection boxes. This is because the detected bounding boxes inherently contain complete geospatial information, aligning more closely with the training paradigm of the CLIP model. As a result, the extracted features are more compact and exhibit higher correspondence with textual descriptions, whereas the randomly cropped regions from MSAR are less conducive to concentrated textual information. In summary, our proposed SRR-TA method not only addresses the long-tail problem but also better facilitates the enhancement of textual features.

\subsection{Which Features are the Most Critical?}

In this section, we analysis the impact of pre-trained VLMs features on UHR satellite image performance.

It is widely acknowledged that visual-linguistic models based on the CLIP\cite{CLIP} architecture consist of at least two branches: a text branch designed to extract text features and an image branch intended to extract visual features. This naturally raises the question of whether the text features extracted by the text branch and the visual features extracted by the image branch both contribute to enhancing the performance of our task. To explore this question, we conducted the following experiments.

\begin{table*}[tbp]
\centering
\caption{The results on the FBP dataset across three experiments.}
\label{tab:datasample_fbp}
\begin{tabularx}{\textwidth}{lcccc ccccc cccc}
\toprule
\textbf{Method} & \textbf{Backbone} & \textbf{IA} & \textbf{PF} & \textbf{IF} & \textbf{DC} & \textbf{GL} & \textbf{AF} & \textbf{SF} & \textbf{PK} & \textbf{NM} & \textbf{AM} & \textbf{RV} & \textbf{UR} \\
\midrule
\multicolumn{14}{c}{\textbf{Investigation of MLMC Cropping and GSD-preserving Method}} \\
\midrule
MSAR \& GSD-preserving & MIT-B5 & 64.13 & 59.95 & 77.46 & 45.02 & 32.11 & 82.18 & 20.92 & 30.57 & 58.33 & 41.61 & 75.63 & 73.58\\
MSAR \& w/o GSD-preserving & MIT-B5 & 52.46 & 47.37 & 70.33 & 4.83 & 18.17 & 78.24 & 5.47 & 9.53 & 55.03 & 12.61 & 74.62 & 65.79\\
WG-ResCro & MIT-B5 & 63.61 & 61.87 & 76.64 & 42.36 & 27.40 & 82.25 & 17.33 & 28.21 & 57.35 & 30.97 & 74.28 & 72.81\\
\midrule
\multicolumn{14}{c}{\textbf{The Impact of Pre-Trained VLMs Features}} \\
\midrule
Txt Feature & MIT-B5 & 64.55 & 60.98 & 77.04 & 44.39 & 27.18 & 82.41 & 19.83 & 28.38 & 57.72 & 40.01 & 75.69 & 73.65\\
Img Feature & MIT-B5 & 63.36 & 58.69 & 77.36 & 40.28 & 30.21 & 82.21 & 23.42 & 25.49 & 55.67 & 38.98 & 72.59 & 72.44\\
Txt-Img Feature & MIT-B5 & 61.28 & 58.16 & 74.77 & 18.90 & 25.41 & 81.25 & 18.98 & 34.25 & 59.87 & 33.62 & 69.27 & 71.13\\
\bottomrule
\end{tabularx}

\begin{tabularx}{\textwidth}{l ccccc ccccc ccc}
\toprule
\textbf{Method} & \textbf{LK} & \textbf{PN} & \textbf{FP} & \textbf{SN} & \textbf{BL} & \textbf{RR} & \textbf{ST} & \textbf{SQ} & \textbf{RO} & \textbf{OP} & \textbf{RS} & \textbf{AP} & \textbf{mIoU(\%)} \\
\midrule
\multicolumn{14}{c}{\textbf{Investigation of MSAR and GSD-preserving Method}} \\
\midrule
MSAR \& GSD-preserving & 82.84 & 25.87 & 63.07 & 12.73 & 43.76 & 64.36 & 39.11 & 16.29 & 62.74 & 63.47 & 45.54 & 49.29 & 51.27  \\
MSAR \& w/o GSD-preserving & 83.58 & 12.04 & 36.38 & 0.00 & 22.94 & 56.85 & 19.63 & 1.06 & 42.06 & 23.68 & 21.02 & 25.03 & 37.89  \\
WG-ResCro & 82.03 & 18.52 & 68.87 & 0.96 & 45.39 & 64.11 & 44.2 & 19.44 & 61.52 & 63.93 & 42.53 & 65.16 & 48.17  \\
\midrule
\multicolumn{14}{c}{\textbf{The Impact of Pre-Trained VLMs Features}} \\
\midrule
TxT Feature & 82.79 & 25.68 & 68.42 & 11.05 & 45.53 & 64.90 & 46.26 & 16.91 & 62.81 & 64.45 & 45.21 & 54.54 & 51.68  \\
Img Feature & 81.37 & 25.63 & 63.84 & 12.12 & 43.52 & 63.34 & 42.69 & 19.19 & 62.56 & 63.84 & 42.91 & 55.84 & 50.73  \\
Txt-Img Feature & 80.33 & 27.92 & 64.19 & 5.67 & 40.69 & 62.76 & 41.42 & 17.84 & 58.39 & 62.31 & 44.84 & 52.92 & 48.59  \\
\bottomrule
\end{tabularx}
\footnotesize
\begin{tablenotes}
    \item \textbf{Abbreviations:} IA = Industrial Area, PF = Paddy Field, IF = Irrigated Field, DC = Dry Cropland, GL = Garden Land, AF = Arbor Forest, SF = Shrub Forest, PK = Park, NM = Natural Meadow, AM = Artificial Meadow, RV = River, UR = Urban Residential, LK = Lake, PN = Pond, FP = Fish Pond, SN = Snow, BL = Bareland, RR = Rural Residential, ST = Stadium, SQ = Square, RD = Road, OP = Overpass, RS = Railway Station, AP = Airport.
\end{tablenotes}
\end{table*}

\subsubsection{Experiment with Feature Extraction Variation} 

We maintained the method of feature infusion as described in Section \ref{subsec:inject}, altering only the features extracted. The text features were still based on our settings in Section \ref{subsec:inject}. To leverage the image branch of the CLIP model, we employed RSSD\cite{GeoRSCLIP} image generation model trained with remote sensing-specific data from our previous work. Using an empty template, we generated five images for each remote sensing-specific land cover class described in Section \ref{subsec:inject}, serving as input for a universal representation. We employed GeoRSCLIP\cite{GeoRSCLIP} to extract visual features from each generated image and averaged the features of the five images for each class. By stacking the averaged features of all classes, we obtained visual features of the same shape as the text features, which were then used to replace the text features infused into the model.

\subsubsection{Further Experiment with Multi-modal Feature Fusion}

Subsequently, we normalized the image and text features separately and concatenated them to create a multi-modal feature containing richer information. This feature set, encompassing a broader spectrum of data, was used to replace the text features infused into the model as a comparative experiment. 

\subsubsection{Analysis of Experimental Results}

The experimental outcomes for the GID, URUR, and FBP datasets are respectively presented in Tables \ref{tab:datasample_gid}, \ref{tab:datasample_urur}, and \ref{tab:datasample_fbp}, under the section titled \textbf{The Impact of Pre-Trained VLMs Features}. It is evident that the information from the text modality alone, while retaining the foundations of MSAR and GSD-preserving, has further enhanced performance. In contrast, both the image unimodal and multimodal information have diminished the model's semantic segmentation accuracy. We proceed to analyze this from a qualitative perspective.

One contributing factor is the inherent differences between the image and text modalities when describing the same object. Text semantics offer a broad latitude for interpretation, with substantial implicit information. During the training process of the CLIP model, the text templates are short, such as \textbf{an image of {}} Even though the length of text is expanded in remote sensing downstream tasks, the effective length remains within 20-100 words. This training approach endows CLIP-based multimodal models with high semantic diversity, enabling the conveyance of a broader range of implicit meanings through minimal text. In contrast, the image modality has minimal semantic latitude, with concrete and limited information. Its features are confined within the boundaries of a single image, exhibiting poor scalability, and as raw input for a universal representation, it lacks the capability to aggregate information.

\begin{figure}[!t]
\centering
\includegraphics[width=\columnwidth]{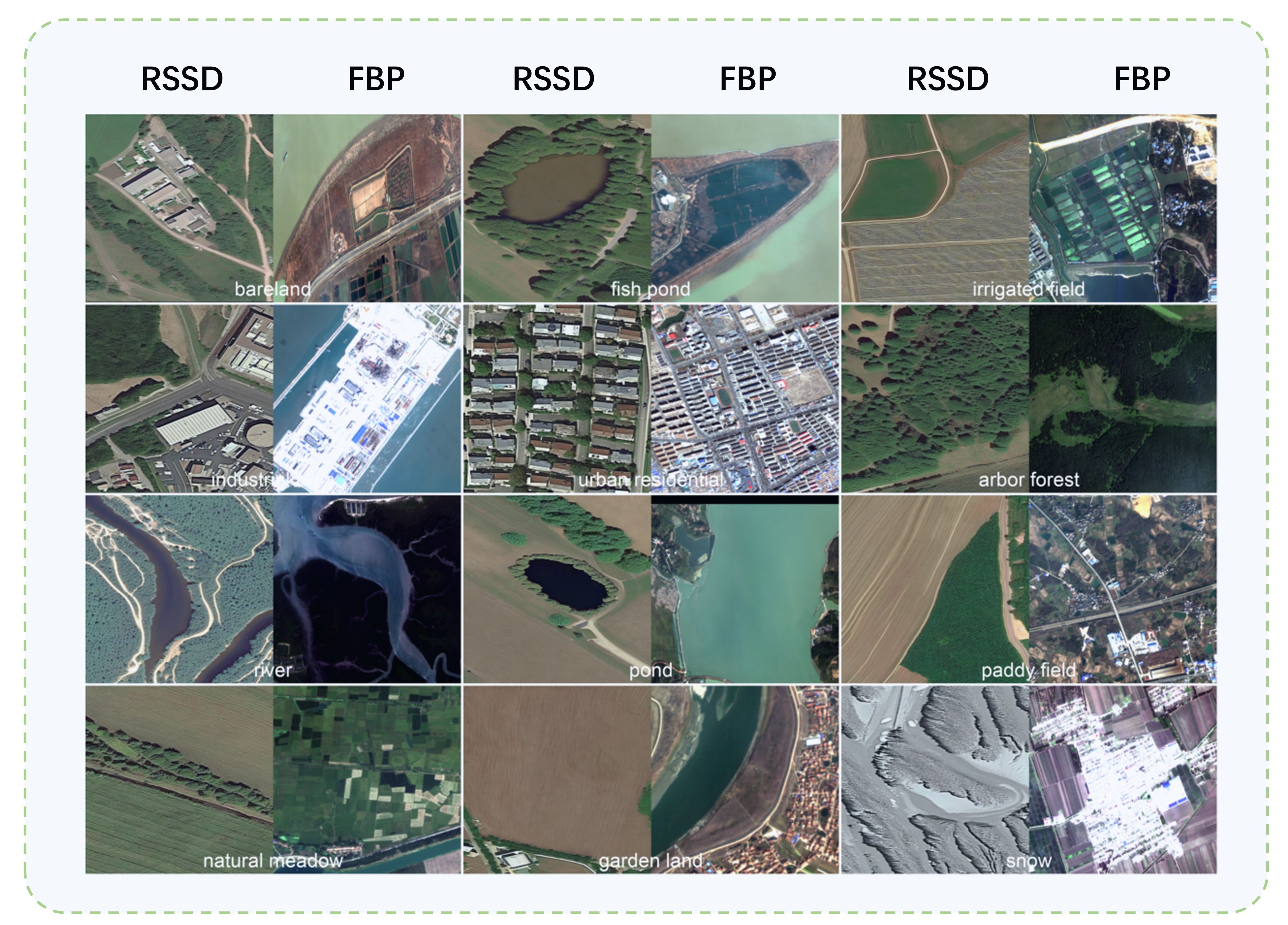}
\caption{Comparison of Images Generated by RSSD and Sampled Images from the FBP Dataset. We sampled 12 categories to compare the differences between the generated images and the authentic remote sensing images within the FBP dataset.}
\label{fig:sd_fbp}
\end{figure}

Another factor is the distinct perspective difference between images generated by Stable Diffusion and actual remote sensing images, as illustrated in Figure \ref{fig:sd_fbp}. It can be observed from the figures that RSSD tends to generate singular objects in response to descriptions, with a relatively uniform and change-lacking environment surrounding these objects. Moreover, the images generated by RSSD\cite{GeoRSCLIP} often resemble aerial photography in terms of elevation and angle, which significantly differs from satellite imagery. In fact, the cosine similarity between the generated images and the original text in GeoRSCLIP\cite{GeoRSCLIP} is below 0.3. This lack of variability in the depiction of single objects and the differences in shooting height and angle mean that using the generated images as raw input for a universal representation does not contribute to the advancement of our task.

\section{Conclusion And Future Work}
\label{sec:conclusion}

In this study, we have revisited the task of UHR image semantic segmentation from the perspective of long-tail distribution, attempting to introduce methods to address this issue within the field. Consequently, we proposed the SRMF framework, which incorporates two data augmentation techniques, MSAR (Multi-Scale Anchored Region Sampling) and SRR-TA (Semantic Reranking and Resampling for Training Augmentation). By integrating text information for the first time without text annotations for individual images, we fused text features with visual features for pixel-level classification, achieving state-of-the-art (SOTA) results in the open-source domain. Our approach not only offers a solution to the long-tail problem in the UHR image segmentation domain but also demonstrates the immense potential of multimodal feature fusion in this area. Our method achieves an improvement of 3.33\%, 0.66\%, and 0.98\% in mIoU on the URUR, GID, and FBP datasets, respectively, attained the SOTA in the field of UHR image semantic segmentation.

Certainly, our method still has several limitations. First, the proposed SRR-TA method heavily relies on SAM-HQ\cite{samhq} detection results. If tail categories are absent in the detected regions, the corresponding class performance will degrade significantly. However, the class distribution of specific data cannot be determined before training, making the efficient pre-extraction of long-tail targets a persistent challenge. Second, our Injection of General Representation Knowledge approach requires a sufficiently comprehensive collection of remote sensing land cover terms in advance. If unseen land cover categories appear in the test data that were not included in the collected vocabulary, the segmentation performance for those categories is likely to decline.

Looking forward, our subsequent research will continue to focus on more effective multimodal feature learning, including the integration of additional modalities, the development of more efficient and powerful feature fusion modules. For instance, integrating various infrared channels, elevation data, location, and temporal information enhances the functionality. Additionally, further exploration of the long-tail problem in remote sensing image processing remains a key focus of our future work. These efforts aim to provide more practical solutions for UHR image segmentation.

\section{ACKNOWLEDGMENT}

This study received funding from the National Key Research and Development Program of China under grant number 2023YFD2000101 and the Hainan Province Science and Technology Special Fund of the Hainan Provincial Department of Science and Technology (ZDYF2022SHFZ323).

\bibliographystyle{IEEEtran}
\bibliography{ref}

\begin{thebibliography}{10}
\providecommand{\url}[1]{#1}
\csname url@samestyle\endcsname
\providecommand{\newblock}{\relax}
\providecommand{\bibinfo}[2]{#2}
\providecommand{\BIBentrySTDinterwordspacing}{\spaceskip=0pt\relax}
\providecommand{\BIBentryALTinterwordstretchfactor}{4}
\providecommand{\BIBentryALTinterwordspacing}{\spaceskip=\fontdimen2\font plus
\BIBentryALTinterwordstretchfactor\fontdimen3\font minus \fontdimen4\font\relax}
\providecommand{\BIBforeignlanguage}[2]{{%
\expandafter\ifx\csname l@#1\endcsname\relax
\typeout{** WARNING: IEEEtran.bst: No hyphenation pattern has been}%
\typeout{** loaded for the language `#1'. Using the pattern for}%
\typeout{** the default language instead.}%
\else
\language=\csname l@#1\endcsname
\fi
#2}}
\providecommand{\BIBdecl}{\relax}
\BIBdecl

\bibitem{lian2020road}
R.~Lian, W.~Wang, N.~Mustafa, and L.~Huang, ``Road extraction methods in high-resolution remote sensing images: A comprehensive review,'' \emph{IEEE Journal of Selected Topics in Applied Earth Observations and Remote Sensing}, vol.~13, pp. 5489--5507, 2020.

\bibitem{Glh-water}
Y.~Li, B.~Dang, W.~Li, and Y.~Zhang, ``Glh-water: A large-scale dataset for global surface water detection in large-size very-high-resolution satellite imagery,'' in \emph{Proceedings of the AAAI Conference on Artificial Intelligence}, vol.~38, no.~20, 2024, pp. 22\,213--22\,221.

\bibitem{brrnet}
Z.~Shao, P.~Tang, Z.~Wang, N.~Saleem, S.~Yam, and C.~Sommai, ``Brrnet: A fully convolutional neural network for automatic building extraction from high-resolution remote sensing images,'' \emph{Remote Sensing}, vol.~12, no.~6, p. 1050, 2020.

\bibitem{luo2024multimodal}
H.~Luo, X.~Feng, B.~Du, and Y.~Zhang, ``A multimodal feature fusion network for building extraction with very high-resolution remote sensing image and lidar data,'' \emph{IEEE Transactions on Geoscience and Remote Sensing}, vol.~62, pp. 1--19, 2024.

\bibitem{forest}
A.~M. Lechner, G.~M. Foody, and D.~S. Boyd, ``Applications in remote sensing to forest ecology and management,'' \emph{One Earth}, vol.~2, no.~5, pp. 405--412, 2020.

\bibitem{lclu}
J.~Rogan and D.~Chen, ``Remote sensing technology for mapping and monitoring land-cover and land-use change,'' \emph{Progress in planning}, vol.~61, no.~4, pp. 301--325, 2004.

\bibitem{shafapourtehrany2023comprehensive}
M.~Shafapourtehrany, M.~Batur, F.~Shabani, B.~Pradhan, B.~Kalantar, and H.~{\"O}zener, ``A comprehensive review of geospatial technology applications in earthquake preparedness, emergency management, and damage assessment,'' \emph{Remote Sensing}, vol.~15, no.~7, p. 1939, 2023.

\bibitem{himeur2022using}
Y.~Himeur, B.~Rimal, A.~Tiwary, and A.~Amira, ``Using artificial intelligence and data fusion for environmental monitoring: A review and future perspectives,'' \emph{Information Fusion}, vol.~86, pp. 44--75, 2022.

\bibitem{li2020review}
J.~Li, Y.~Pei, S.~Zhao, R.~Xiao, X.~Sang, and C.~Zhang, ``A review of remote sensing for environmental monitoring in china,'' \emph{Remote Sensing}, vol.~12, no.~7, p. 1130, 2020.

\bibitem{radovcaj2020global}
D.~Rado{\v{c}}aj, J.~Obho{\dj}a{\v{s}}, M.~Juri{\v{s}}i{\'c}, and M.~Ga{\v{s}}parovi{\'c}, ``Global open data remote sensing satellite missions for land monitoring and conservation: A review,'' \emph{Land}, vol.~9, no.~11, p. 402, 2020.

\bibitem{lin2021identifying}
A.~Lin, X.~Sun, H.~Wu, W.~Luo, D.~Wang, D.~Zhong, Z.~Wang, L.~Zhao, and J.~Zhu, ``Identifying urban building function by integrating remote sensing imagery and poi data,'' \emph{IEEE Journal of Selected Topics in Applied Earth Observations and Remote Sensing}, vol.~14, pp. 8864--8875, 2021.

\bibitem{zhu2021detecting}
D.~Zhu, T.~Chen, Z.~Wang, and R.~Niu, ``Detecting ecological spatial-temporal changes by remote sensing ecological index with local adaptability,'' \emph{Journal of Environmental Management}, vol. 299, p. 113655, 2021.

\bibitem{iterdanet}
Y.~Cai, Y.~Yang, Y.~Shang, Z.~Chen, Z.~Shen, and J.~Yin, ``Iterdanet: Iterative intra-domain adaptation for semantic segmentation of remote sensing images,'' \emph{IEEE Transactions on Geoscience and Remote Sensing}, vol.~60, pp. 1--17, 2022.

\bibitem{bifdanet}
Y.~Cai, Y.~Yang, Q.~Zheng, Z.~Shen, Y.~Shang, J.~Yin, and Z.~Shi, ``Bifdanet: Unsupervised bidirectional domain adaptation for semantic segmentation of remote sensing images,'' \emph{Remote Sensing}, vol.~14, no.~1, p. 190, 2022.

\bibitem{dbblendmask}
Z.~Chen, Y.~Shang, A.~Python, Y.~Cai, and J.~Yin, ``Db-blendmask: Decomposed attention and balanced blendmask for instance segmentation of high-resolution remote sensing images,'' \emph{IEEE Transactions on Geoscience and Remote Sensing}, vol.~60, pp. 1--15, 2022.

\bibitem{deeplabv3p}
L.-C. Chen, Y.~Zhu, G.~Papandreou, F.~Schroff, and H.~Adam, ``Encoder-decoder with atrous separable convolution for semantic image segmentation,'' in \emph{Proceedings of the European conference on computer vision (ECCV)}, 2018, pp. 801--818.

\bibitem{fpn}
A.~Kirillov, R.~Girshick, K.~He, and P.~Doll{\'a}r, ``Panoptic feature pyramid networks,'' in \emph{Proceedings of the IEEE/CVF conference on computer vision and pattern recognition}, 2019, pp. 6399--6408.

\bibitem{pspnet}
H.~Zhao, J.~Shi, X.~Qi, X.~Wang, and J.~Jia, ``Pyramid scene parsing network,'' in \emph{Proceedings of the IEEE conference on computer vision and pattern recognition}, 2017, pp. 2881--2890.

\bibitem{unet}
O.~Ronneberger, P.~Fischer, and T.~Brox, ``U-net: Convolutional networks for biomedical image segmentation,'' in \emph{Medical image computing and computer-assisted intervention--MICCAI 2015: 18th international conference, Munich, Germany, October 5-9, 2015, proceedings, part III 18}.\hskip 1em plus 0.5em minus 0.4em\relax Springer, 2015, pp. 234--241.

\bibitem{unet++}
Z.~Zhou, M.~M. Rahman~Siddiquee, N.~Tajbakhsh, and J.~Liang, ``Unet++: A nested u-net architecture for medical image segmentation,'' in \emph{Deep Learning in Medical Image Analysis and Multimodal Learning for Clinical Decision Support: 4th International Workshop, DLMIA 2018, and 8th International Workshop, ML-CDS 2018, Held in Conjunction with MICCAI 2018, Granada, Spain, September 20, 2018, Proceedings 4}.\hskip 1em plus 0.5em minus 0.4em\relax Springer, 2018, pp. 3--11.

\bibitem{segformer}
E.~Xie, W.~Wang, Z.~Yu, A.~Anandkumar, J.~M. Alvarez, and P.~Luo, ``Segformer: Simple and efficient design for semantic segmentation with transformers,'' \emph{Advances in neural information processing systems}, vol.~34, pp. 12\,077--12\,090, 2021.

\bibitem{pvtv2}
W.~Wang, E.~Xie, X.~Li, D.-P. Fan, K.~Song, D.~Liang, T.~Lu, P.~Luo, and L.~Shao, ``Pvt v2: Improved baselines with pyramid vision transformer,'' \emph{Computational Visual Media}, vol.~8, no.~3, pp. 415--424, 2022.

\bibitem{strudel2021segmenter}
R.~Strudel, R.~Garcia, I.~Laptev, and C.~Schmid, ``Segmenter: Transformer for semantic segmentation,'' in \emph{Proceedings of the IEEE/CVF international conference on computer vision}, 2021, pp. 7262--7272.

\bibitem{vit}
A.~Dosovitskiy, L.~Beyer, A.~Kolesnikov, D.~Weissenborn, X.~Zhai, T.~Unterthiner, M.~Dehghani, M.~Minderer, G.~Heigold, S.~Gelly \emph{et~al.}, ``An image is worth 16x16 words: Transformers for image recognition at scale,'' \emph{arXiv preprint arXiv:2010.11929}, 2020.

\bibitem{swintrans}
Z.~Liu, Y.~Lin, Y.~Cao, H.~Hu, Y.~Wei, Z.~Zhang, S.~Lin, and B.~Guo, ``Swin transformer: Hierarchical vision transformer using shifted windows,'' in \emph{Proceedings of the IEEE/CVF international conference on computer vision}, 2021, pp. 10\,012--10\,022.

\bibitem{GLNet}
W.~Chen, Z.~Jiang, Z.~Wang, K.~Cui, and X.~Qian, ``Collaborative global-local networks for memory-efficient segmentation of ultra-high resolution images,'' in \emph{Proceedings of the IEEE/CVF conference on computer vision and pattern recognition}, 2019, pp. 8924--8933.

\bibitem{wconet}
L.~Ding, D.~Lin, S.~Lin, J.~Zhang, X.~Cui, Y.~Wang, H.~Tang, and L.~Bruzzone, ``Looking outside the window: Wide-context transformer for the semantic segmentation of high-resolution remote sensing images,'' \emph{IEEE Transactions on Geoscience and Remote Sensing}, vol.~60, pp. 1--13, 2022.

\bibitem{icnet}
H.~Zhao, X.~Qi, X.~Shen, J.~Shi, and J.~Jia, ``Icnet for real-time semantic segmentation on high-resolution images,'' in \emph{Proceedings of the European conference on computer vision (ECCV)}, 2018, pp. 405--420.

\bibitem{Fctlnet}
Q.~Li, W.~Yang, W.~Liu, Y.~Yu, and S.~He, ``From contexts to locality: Ultra-high resolution image segmentation via locality-aware contextual correlation,'' in \emph{Proceedings of the IEEE/CVF International Conference on Computer Vision}, 2021, pp. 7252--7261.

\bibitem{isdnet}
S.~Guo, L.~Liu, Z.~Gan, Y.~Wang, W.~Zhang, C.~Wang, G.~Jiang, W.~Zhang, R.~Yi, L.~Ma \emph{et~al.}, ``Isdnet: Integrating shallow and deep networks for efficient ultra-high resolution segmentation,'' in \emph{Proceedings of the IEEE/CVF Conference on Computer Vision and Pattern Recognition}, 2022, pp. 4361--4370.

\bibitem{ehsnet}
W.~Chen, Y.~Li, B.~Dang, and Y.~Zhang, ``Ehsnet: End-to-end holistic learning network for large-size remote sensing image semantic segmentation,'' \emph{arXiv preprint arXiv:2211.11316}, 2022.

\bibitem{urur}
D.~Ji, F.~Zhao, H.~Lu, M.~Tao, and J.~Ye, ``Ultra-high resolution segmentation with ultra-rich context: A novel benchmark,'' in \emph{Proceedings of the IEEE/CVF Conference on Computer Vision and Pattern Recognition}, 2023, pp. 23\,621--23\,630.

\bibitem{GPWFormer}
D.~Ji, F.~Zhao, and H.~Lu, ``Guided patch-grouping wavelet transformer with spatial congruence for ultra-high resolution segmentation,'' \emph{arXiv preprint arXiv:2307.00711}, 2023.

\bibitem{patchproposal}
T.~Wu, Z.~Lei, B.~Lin, C.~Li, Y.~Qu, and Y.~Xie, ``Patch proposal network for fast semantic segmentation of high-resolution images,'' in \emph{Proceedings of the AAAI Conference on Artificial Intelligence}, vol.~34, no.~07, 2020, pp. 12\,402--12\,409.

\bibitem{GeoAgent}
Y.~Liu, S.~Shi, J.~Wang, and Y.~Zhong, ``Seeing beyond the patch: Scale-adaptive semantic segmentation of high-resolution remote sensing imagery based on reinforcement learning,'' in \emph{Proceedings of the IEEE/CVF International Conference on Computer Vision}, 2023, pp. 16\,868--16\,878.

\bibitem{FBP}
X.-Y. Tong, G.-S. Xia, and X.~X. Zhu, ``Enabling country-scale land cover mapping with meter-resolution satellite imagery,'' \emph{ISPRS Journal of Photogrammetry and Remote Sensing}, vol. 196, pp. 178--196, 2023.

\bibitem{ssrs_survey}
L.~Huang, B.~Jiang, S.~Lv, Y.~Liu, and Y.~Fu, ``Deep-learning-based semantic segmentation of remote sensing images: A survey,'' \emph{IEEE Journal of Selected Topics in Applied Earth Observations and Remote Sensing}, vol.~17, pp. 8370--8396, 2023.

\bibitem{lt-93}
R.~Anand, K.~G. Mehrotra, C.~K. Mohan, and S.~Ranka, ``An improved algorithm for neural network classification of imbalanced training sets,'' \emph{IEEE transactions on neural networks}, vol.~4, no.~6, pp. 962--969, 1993.

\bibitem{lt-nc}
Z.~Zhong, J.~Cui, Y.~Yang, X.~Wu, X.~Qi, X.~Zhang, and J.~Jia, ``Understanding imbalanced semantic segmentation through neural collapse,'' in \emph{Proceedings of the IEEE/CVF conference on computer vision and pattern recognition}, 2023, pp. 19\,550--19\,560.

\bibitem{lt-ss}
J.~Cui, Y.~Yuan, Z.~Zhong, Z.~Tian, H.~Hu, S.~Lin, and J.~Jia, ``Region rebalance for long-tailed semantic segmentation,'' \emph{arXiv preprint arXiv:2204.01969}, 2022.

\bibitem{oversample}
H.~Han, W.-Y. Wang, and B.-H. Mao, ``Borderline-smote: a new over-sampling method in imbalanced data sets learning,'' in \emph{International conference on intelligent computing}.\hskip 1em plus 0.5em minus 0.4em\relax Springer, 2005, pp. 878--887.

\bibitem{undersample}
C.~Drummond, R.~C. Holte \emph{et~al.}, ``C4. 5, class imbalance, and cost sensitivity: why under-sampling beats over-sampling,'' in \emph{Workshop on learning from imbalanced datasets II}, vol.~11, no. 1--8, 2003.

\bibitem{balancesample1}
D.~Mahajan, R.~Girshick, V.~Ramanathan, K.~He, M.~Paluri, Y.~Li, A.~Bharambe, and L.~Van Der~Maaten, ``Exploring the limits of weakly supervised pretraining,'' in \emph{Proceedings of the European conference on computer vision (ECCV)}, 2018, pp. 181--196.

\bibitem{balancesample2}
L.~Shen, Z.~Lin, and Q.~Huang, ``Relay backpropagation for effective learning of deep convolutional neural networks,'' in \emph{Computer Vision--ECCV 2016: 14th European Conference, Amsterdam, The Netherlands, October 11--14, 2016, Proceedings, Part VII 14}.\hskip 1em plus 0.5em minus 0.4em\relax Springer, 2016, pp. 467--482.

\bibitem{sam}
A.~Kirillov, E.~Mintun, N.~Ravi, H.~Mao, C.~Rolland, L.~Gustafson, T.~Xiao, S.~Whitehead, A.~C. Berg, W.-Y. Lo \emph{et~al.}, ``Segment anything,'' in \emph{Proceedings of the IEEE/CVF International Conference on Computer Vision}, 2023, pp. 4015--4026.

\bibitem{samhq}
L.~Ke, M.~Ye, M.~Danelljan, Y.-W. Tai, C.-K. Tang, F.~Yu \emph{et~al.}, ``Segment anything in high quality,'' \emph{Advances in Neural Information Processing Systems}, vol.~36, 2024.

\bibitem{protoseg}
T.~Zhou, W.~Wang, E.~Konukoglu, and L.~Van~Gool, ``Rethinking semantic segmentation: A prototype view,'' in \emph{Proceedings of the IEEE/CVF Conference on Computer Vision and Pattern Recognition}, 2022, pp. 2582--2593.

\bibitem{wavelet}
D.~Ji, F.~Zhao, and H.~Lu, ``Guided patch-grouping wavelet transformer with spatial congruence for ultra-high resolution segmentation,'' \emph{arXiv preprint arXiv:2307.00711}, 2023.

\bibitem{CLIP}
A.~Radford, J.~W. Kim, C.~Hallacy, A.~Ramesh, G.~Goh, S.~Agarwal, G.~Sastry, A.~Askell, P.~Mishkin, J.~Clark \emph{et~al.}, ``Learning transferable visual models from natural language supervision,'' in \emph{International conference on machine learning}.\hskip 1em plus 0.5em minus 0.4em\relax PMLR, 2021, pp. 8748--8763.

\bibitem{GeoRSCLIP}
Z.~Zhang, T.~Zhao, Y.~Guo, and J.~Yin, ``Rs5m: A large scale vision-language dataset for remote sensing vision-language foundation model,'' \emph{arXiv preprint arXiv:2306.11300}, 2023.

\bibitem{RemoteCLIP}
F.~Liu, D.~Chen, Z.~Guan, X.~Zhou, J.~Zhu, Q.~Ye, L.~Fu, and J.~Zhou, ``Remoteclip: A vision language foundation model for remote sensing,'' \emph{IEEE Transactions on Geoscience and Remote Sensing}, 2024.

\bibitem{rsclip}
X.~Li, C.~Wen, Y.~Hu, and N.~Zhou, ``Rs-clip: Zero shot remote sensing scene classification via contrastive vision-language supervision,'' \emph{International Journal of Applied Earth Observation and Geoinformation}, vol. 124, p. 103497, 2023.

\bibitem{semantic_sam}
F.~Li, H.~Zhang, P.~Sun, X.~Zou, S.~Liu, J.~Yang, C.~Li, L.~Zhang, and J.~Gao, ``Semantic-sam: Segment and recognize anything at any granularity,'' \emph{arXiv preprint arXiv:2307.04767}, 2023.

\bibitem{presam}
R.~Zhang, Z.~Jiang, Z.~Guo, S.~Yan, J.~Pan, X.~Ma, H.~Dong, P.~Gao, and H.~Li, ``Personalize segment anything model with one shot,'' \emph{arXiv preprint arXiv:2305.03048}, 2023.

\bibitem{sam_changedetect}
L.~Ding, K.~Zhu, D.~Peng, H.~Tang, K.~Yang, and L.~Bruzzone, ``Adapting segment anything model for change detection in vhr remote sensing images,'' \emph{IEEE Transactions on Geoscience and Remote Sensing}, 2024.

\bibitem{sam_changeosm}
H.~Chen, J.~Song, and N.~Yokoya, ``Change detection between optical remote sensing imagery and map data via segment anything model (sam),'' \emph{arXiv preprint arXiv:2401.09019}, 2024.

\bibitem{sam_evaluating}
B.~Gui, A.~Bhardwaj, and L.~Sam, ``Evaluating the efficacy of segment anything model for delineating agriculture and urban green spaces in multiresolution aerial and spaceborne remote sensing images,'' \emph{Remote sensing}, vol.~16, no.~2, p. 414, 2024.

\bibitem{mesam}
X.~Zhou, F.~Liang, L.~Chen, H.~Liu, Q.~Song, G.~Vivone, and J.~Chanussot, ``Mesam: Multiscale enhanced segment anything model for optical remote sensing images,'' \emph{IEEE Transactions on Geoscience and Remote Sensing}, 2024.

\bibitem{sam4rs}
L.~P. Osco, Q.~Wu, E.~L. de~Lemos, W.~N. Gon{\c{c}}alves, A.~P.~M. Ramos, J.~Li, and J.~M. Junior, ``The segment anything model (sam) for remote sensing applications: From zero to one shot,'' \emph{International Journal of Applied Earth Observation and Geoinformation}, vol. 124, p. 103540, 2023.

\bibitem{lt-or}
J.~Tan, C.~Wang, B.~Li, Q.~Li, W.~Ouyang, C.~Yin, and J.~Yan, ``Equalization loss for long-tailed object recognition,'' in \emph{Proceedings of the IEEE/CVF conference on computer vision and pattern recognition}, 2020, pp. 11\,662--11\,671.

\bibitem{GID}
X.-Y. Tong, G.-S. Xia, Q.~Lu, H.~Shen, S.~Li, S.~You, and L.~Zhang, ``Land-cover classification with high-resolution remote sensing images using transferable deep models,'' \emph{Remote Sensing of Environment}, vol. 237, p. 111322, 2020.

\bibitem{whuopt}
X.~Li, G.~Zhang, H.~Cui, S.~Hou, S.~Wang, X.~Li, Y.~Chen, Z.~Li, and L.~Zhang, ``Mcanet: A joint semantic segmentation framework of optical and sar images for land use classification,'' \emph{International Journal of Applied Earth Observation and Geoinformation}, vol. 106, p. 102638, 2022.

\bibitem{deepglobe}
I.~Demir, K.~Koperski, D.~Lindenbaum, G.~Pang, J.~Huang, S.~Basu, F.~Hughes, D.~Tuia, and R.~Raskar, ``Deepglobe 2018: A challenge to parse the earth through satellite images,'' in \emph{Proceedings of the IEEE conference on computer vision and pattern recognition workshops}, 2018, pp. 172--181.

\bibitem{mmseg}
M.~Contributors, ``{MMSegmentation}: Openmmlab semantic segmentation toolbox and benchmark,'' \url{https://github.com/open-mmlab/mmsegmentation}, 2020.

\bibitem{upernet}
T.~Xiao, Y.~Liu, B.~Zhou, Y.~Jiang, and J.~Sun, ``Unified perceptual parsing for scene understanding,'' in \emph{Proceedings of the European conference on computer vision (ECCV)}, 2018, pp. 418--434.

\bibitem{fcn}
J.~Long, E.~Shelhamer, and T.~Darrell, ``Fully convolutional networks for semantic segmentation,'' in \emph{Proceedings of the IEEE conference on computer vision and pattern recognition}, 2015, pp. 3431--3440.

\bibitem{bedsn}
X.~Li, L.~Xie, C.~Wang, J.~Miao, H.~Shen, and L.~Zhang, ``Boundary-enhanced dual-stream network for semantic segmentation of high-resolution remote sensing images,'' \emph{GIScience \& Remote Sensing}, vol.~61, no.~1, p. 2356355, 2024.

\end{thebibliography}

\vfill

\end{document}